\documentclass{article}

\usepackage[preprint]{corl_2026} % arXiv/preprint version without the CoRL footer.

\usepackage{amsmath,amssymb}
\usepackage{graphicx}
\usepackage{booktabs}
\usepackage{multirow}
\usepackage{array}
\usepackage{xcolor}
\usepackage{enumitem}
\usepackage{placeins}
\usepackage{flafter}
\usepackage{wrapfig}

\newcommand{\OT}{\mathcal{O}_T}
\newcommand{\OS}{\mathcal{O}_S}
\newcommand{\piT}{\pi_T}
\newcommand{\piS}{\pi_S}

\newcommand{\nrows}{N}
\newcommand{\AME}{\textnormal{AME}}
\newcommand{\TAMSE}{\textnormal{\textsc{TA-MSE}}}
\definecolor{soloBlue}{RGB}{0, 114, 200}
\title{\textcolor{soloBlue}{SOLO}: \textcolor{soloBlue}{S}table \textcolor{soloBlue}{O}mni-terrain \textcolor{soloBlue}{L}ong-Horizon Perceptive Humanoid Loc\textcolor{soloBlue}{o}motion%
\texorpdfstring{\,\raisebox{0.4ex}}{}}

\author{
  \normalfont
  Pihai Sun$^{1,2,*}$, Gang Han$^{2,*}$, Jingkai Sun$^{2,4,*}$,
  Jiahao Ma$^{2,5}$, Zeran Su$^{1,2}$, Zelin Tao$^{2}$,
  Peiran Liu$^{2,3}$ \\[-1pt]
  Shuai Shi$^{2}$, Wei Cui$^{2}$, Zifan Wang$^{3}$,
  Jialin Yu$^{2}$, Wen Zhao$^{2}$, Kangning Yin$^{6}$,
  Jiaxu Wang$^{7}$ \\[-1pt]
  Jiahang Cao$^{4}$, Lingfeng Zhang$^{8}$, Hao Cheng$^{3}$,
  Jian Tang$^{2}$, Qiang Zhang$^{1,\dagger}$, Yijie Guo$^{2}$ \\[2pt]
  \normalfont
  $^1$Artificial General Intelligence Institute,
  University of Science and Technology of China \\
  $^2$X-Humanoid \quad
  $^3$The Hong Kong University of Science and Technology (Guangzhou) \\
  $^4$The University of Hong Kong \quad
  $^5$The Australian National University \\
  $^6$Shanghai Jiao Tong University \quad
  $^7$The Chinese University of Hong Kong \quad
  $^8$Tsinghua University \\[2pt]
}

\begin{document}
\maketitle

\vspace{-40pt}
\begin{center}
  % \small
  \normalfont
  $^*$Equal contribution
  \qquad
  $^\dagger$Corresponding author \\
  \href{https://sunpihai-up.github.io/solo/}
       {\texttt{Project Page: sunpihai-up.github.io/solo}}
\end{center}
% \vspace{-0.8em}

\vspace{-20pt}
\begin{figure}[!hbp]
    \centering
    \includegraphics[width=\columnwidth]{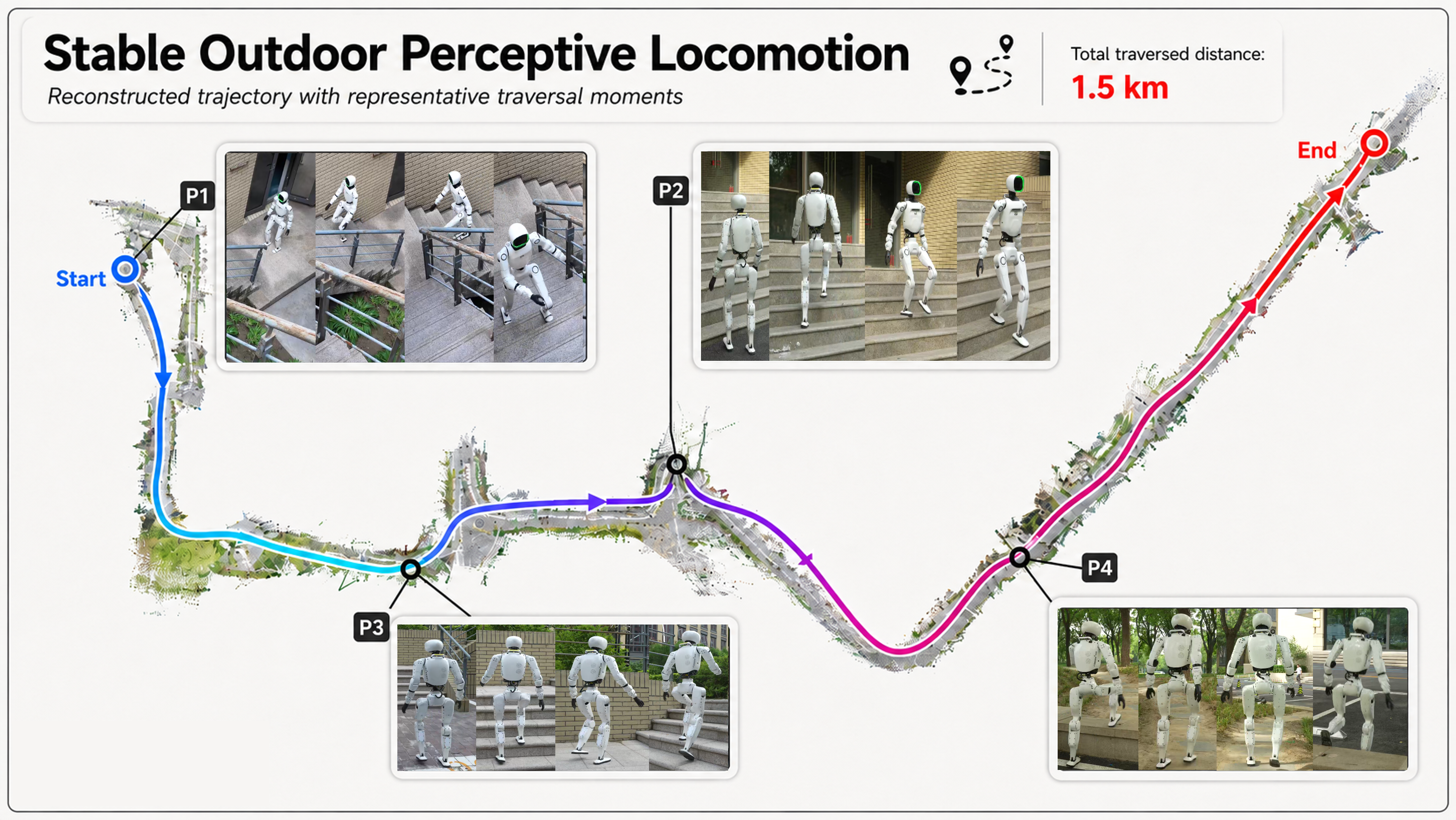}
    \caption{
    \textbf{Stable long-horizon perceptive locomotion in the wild.} SOLO
    completes a continuous \textbf{1.5-km} outdoor route over natural stairs,
    slopes, grass transitions, and uneven ground in one run.
    }
    \label{fig:stable_outdoor_perceptive_locomotion}
\end{figure}

%===============================================================================
\vspace{-10pt}

\begin{abstract}
Humans traverse complex terrain over long distances without losing balance,
whereas perceptive humanoid policies become fragile as perception and control
errors accumulate. We present \textbf{SOLO}, a unified framework addressing
two compounding causes of this long-horizon fragility: dense terrain
reconstruction smooths action-critical details, and pointwise imitation lacks
temporal credit assignment. Its \textbf{Query Reconstructor (QR)} uses
Fourier-encoded cell queries to retrieve spatially specific evidence from
depth--proprioception tokens, preserving sharp terrain boundaries.
\textbf{Trajectory-Aware MSE (TA-MSE) Distillation} adds next-state
teacher--student disagreement to the PPO reward, enabling Generalized
Advantage Estimation to propagate future disagreement penalties to preceding
actions. In simulation, QR reduces height-map L1 error by
$3.3$--$4.0\times$, while TA-MSE surpasses PPO and MSE+PPO in curriculum
progression. On stress-test terrains, SOLO achieves $97.5\%$ mean traversal
success and $96\%$ stepping-stone success, versus $75.0$--$75.6\%$ and
$0$--$3\%$ for dense-reconstructor variants. Deployed zero-shot with only a
chest-mounted depth camera and proprioception, SOLO completes a continuous
\textbf{1.5-km} outdoor route and an indoor mixed-terrain course.
\end{abstract}

%===============================================================================
\section{Introduction}
\label{sec:intro}

Humans traverse trails, uneven stairs, and sparse footholds over long
distances without losing balance.  Their stability couples perception and
control: each foot placement shapes the next state, while corrective actions
prevent errors from causing falls.  Perceptive humanoids operate under
the same feedback loop, yet repeated errors can compound during extended
deployment.  Thus, isolated-obstacle success does not ensure stability over
an uninterrupted route with repeated terrain transitions and viewpoint
changes during real-world deployment throughout sustained operation.

Teacher--student reinforcement learning enables humanoids to traverse
challenging terrain and parkour courses~\citep{radosavovic2024humanoid,
radosavovic2024tokenprediction, cheng2024parkour, hoeller2024parkour,
wang2025beamdojo, zhang2026rpl}.  A privileged simulator teacher is distilled
into a deployable student using onboard depth and proprioception~\citep{
lee2020quadruped, miki2022wild, kumar2021rma, sun2025dpl, yu2026start}.
Short trials reset local errors, whereas continuous deployment repeats terrain
transitions and viewpoint changes that let errors persist and compound.  Two
failures drive this long-horizon fragility in extended real-world rollouts in practice:
\begin{itemize}[leftmargin=*, topsep=2pt, itemsep=2pt]
    \item \textbf{Lossy dense reconstruction.} Standard terrain reconstructors
    compress depth history through CNN--RNN encoders and decode a
    local height map from a shared representation~\citep{
    lee2020quadruped, miki2022wild, cheng2024parkour, sun2025dpl, yu2026start,
    ji2022concurrent}. This shared bottleneck acts as a low-pass filter:
    it captures broad terrain geometry but smooths high-frequency,
    action-critical details such as stepping-stone boundaries and stair edges.
    The loss of per-cell fidelity increases foothold errors that destabilize
    the rollout over long horizons.
    \item \textbf{Myopic pointwise imitation.} Controller
    distillation uses action MSE to match the student to the
    teacher only at the \emph{currently visited} state~\citep{zhang2025dppo,
    cheng2024parkour, zhang2026ame2, rudin2025parkour}. This objective
    cannot distinguish actions with similar current-state error but different
    future consequences. Without trajectory-level credit assignment,
    plausible actions can lead to future states where teacher
    matching becomes difficult.
\end{itemize}

To address these failure modes, SOLO combines two complementary mechanisms.
At the perception level, the \emph{Query Reconstructor (QR)} assigns a
Fourier-encoded query to each height-map cell and cross-attends the queries to
the depth--proprioception token memory. Each query retrieves spatially specific
evidence, preserving sharp terrain boundaries that dense decoders tend to
smooth out. At the policy level, \emph{Trajectory-Aware MSE (TA-MSE)
Distillation} evaluates teacher--student disagreement at the next state and
incorporates it into the PPO reward. Through Generalized Advantage Estimation
(GAE), future disagreement penalties influence preceding actions in the
sampled rollout, providing trajectory-aware credit assignment. At deployment,
SOLO uses onboard depth and proprioception.

We evaluate both components separately and then assess the integrated system.
At the highest curriculum difficulty, QR reduces average height-map L1 error
by $3.3$--$4.0\times$ relative to START- and DPL-style reconstructors.  TA-MSE
reaches higher curriculum levels than PPO and MSE+PPO after both plateau.
With TA-MSE fixed, QR raises mean stress-test success from $75.0$--$75.6\%$ to
$97.5\%$ and stepping-stone success from $0$--$3\%$ to $96\%$.  SOLO transfers
zero-shot to Omni, completing a continuous \textbf{1.5-km} outdoor route and
an indoor mixed-terrain course from onboard depth and proprioception during
real-world deployment using the same onboard sensing stack directly.

The contributions span perception, policy learning, and deployment evidence
across the system:
\begin{itemize}[leftmargin=*, topsep=2pt, itemsep=2pt]
    \item \textbf{Cell-query terrain reconstruction.} We introduce QR, which
    retrieves cell-specific evidence and preserves action-critical terrain
    boundaries from partial depth history reliably during locomotion.
    \item \textbf{Trajectory-aware policy distillation.} We introduce
    TA-MSE, which feeds next-state teacher--student disagreement to PPO credit
    assignment without another critic or teacher rollout, while preserving
    the current-state supervision used in teacher--student training throughout rollouts.
    \item \textbf{Long-horizon evaluation and deployment.} We validate SOLO
    through reconstruction analysis, highest-difficulty simulation stress
    tests, and zero-shot hardware deployment on continuous indoor and outdoor
    routes using only onboard depth and proprioception over extended routes alone.
\end{itemize}
%===============================================================================
\section{Related Work}
\label{sec:related}

\paragraph{Teacher--student perceptive locomotion.}
Privileged learning has become a standard recipe for legged locomotion:
train a teacher with clean terrain and state signals, then distill a student
that uses onboard sensing~\citep{lee2020quadruped, miki2022wild,
cheng2024parkour, zhang2026ame2, rudin2025parkour, kumar2021rma,
agarwal2022egocentric, ji2022concurrent, loquercio2022crossmodal,
yang2022crossmodaltransformer, zhuang2023robotparkour}.  For humanoids,
related work has pushed the same recipe to parkour, stairs, sparse
footholds and other challenging terrains, while broader humanoid
locomotion has scaled from blind walking to robust control
\citep{zhuang2025humanoidparkour, wang2025beamdojo, zhang2026rpl,
radosavovic2024humanoid, radosavovic2024tokenprediction,
li2024humanoidlearningfundamental}.  
Existing systems primarily expand terrain coverage or improve the privileged
teacher; we study student fragility during extended, uninterrupted deployment.
QR and \TAMSE{} target its complementary sources in terrain reconstruction and
policy distillation across long rollouts on extended hardware routes,
respectively, while preserving the same onboard sensing and control interface at run time.

\paragraph{Terrain reconstruction for locomotion.}
Explicit reconstruction has emerged as the dominant way to bridge the sensing
gap between teacher and student.  START~\citep{yu2026start} reconstructs a
local terrain map for sparse foothold traversal with a dense recurrent
decoder, and DPL~\citep{sun2025dpl} couples realistic depth synthesis with a
cross-attention terrain reconstructor for depth-only humanoid locomotion.
Neural volumetric memory and mapping-based pipelines explore alternative
intermediate representations~\citep{hoeller2024parkour, yang2023nvm,
zhang2026ame2, yang2022crossmodaltransformer, loquercio2022crossmodal,
agarwal2022egocentric}. 
QR differs from these reconstructors in its output
parameterization: instead of decoding the height map as a dense grid or a
single shared embedding, each map cell carries its own Fourier-encoded query
that attends to the depth-proprioception memory.  This preserves cell
identity and the small, action-critical terrain features that dense decoders
tend to average away during closed-loop locomotion over narrow, discontinuous footholds
in every evaluated case.

\paragraph{Policy distillation and trajectory-aware imitation.}
Policy distillation transfers teacher behavior through action
distributions, logits or value-aligned objectives~\citep{
hinton2015kd, rusu2015policy, schmitt2018kickstart,
czarnecki2019distilling}.  In locomotion, the practical default is to add an
action MSE auxiliary loss to PPO~\citep{zhang2025dppo, cheng2024parkour,
zhang2026ame2, rudin2025parkour}, which we denote as MSE+PPO.  This single-step objective is easy to
implement but does not model the future states the student itself
induces, a long-standing point of sequential imitation theory~\citep{
ross2011dagger}. 
\TAMSE{} stays close to locomotion practice: it retains the current-state MSE
loss but also inserts next-state teacher--student MSE into the PPO reward, so
rollout credit assignment discourages actions that drive future divergence
from the teacher during long-horizon closed-loop deployment using only onboard
sensing on hardware using the same onboard observations available throughout deployment directly.

\section{Method}
\label{sec:method}

SOLO distills a privileged terrain-conditioned teacher into an onboard
recurrent student.  The Query Reconstructor (QR) estimates the policy's
terrain state, while Trajectory-Aware MSE (\TAMSE{}) uses future
teacher--student disagreement to shape the student-policy updates in Stages~II
and~III.

\subsection{Teacher--Student Setting and System Overview}
\label{sec:problem}

At simulator state $s_t$, the teacher observes
$o_t^T=(m_t,v_t,p_t)$, where $m_t$ is a clean local height map, $v_t$ is
the realized base velocity, and
$p_t=[\omega_t^b,g_t^b,c_t,q_t,\dot q_t,a_{t-1}]\in\mathbb{R}^{84}$
contains proprioception, the planar velocity command $c_t$, and the previous
action.  Thus $c_t$ is distinct from the teacher's realized velocity $v_t$
and QR's estimate $\hat v_t$.  The deployable
student instead receives a history of egocentric depth images and
proprioception,
$o_t^S=(D_{t-T_d+1:t},p_{t-T_p+1:t})$.  QR maps this history to
$(\hat m_t,\hat v_t)$, and the recurrent student combines these estimates
with proprioception to produce an action distribution.  Both policies
output a 25-dimensional vector of offsets from the default joint
positions, which the interface converts to joint targets for execution by the
low-level controller on the real robot at $50$~Hz within the 20-ms onboard control period
on the physical platform itself.

The teacher $\piT$ and student $\piS$ share the Attention-based Map
Encoder (\AME) backbone~\citep{he2025ame1}; the student adds an
LSTM~\citep{hochreiter1997lstm} proprioceptive bypass to retain rollout
history.  Stage~II distills the student with \TAMSE{} while using privileged
terrain input.  Stage~III replaces the privileged terrain state with QR
estimates, jointly fine-tunes the student, and updates QR from the student's
on-policy buffer.  QR supervision follows the states visited by the
QR-conditioned student in Stage~III.

\begin{figure*}[t]
\centering
\includegraphics[width=0.98\textwidth]{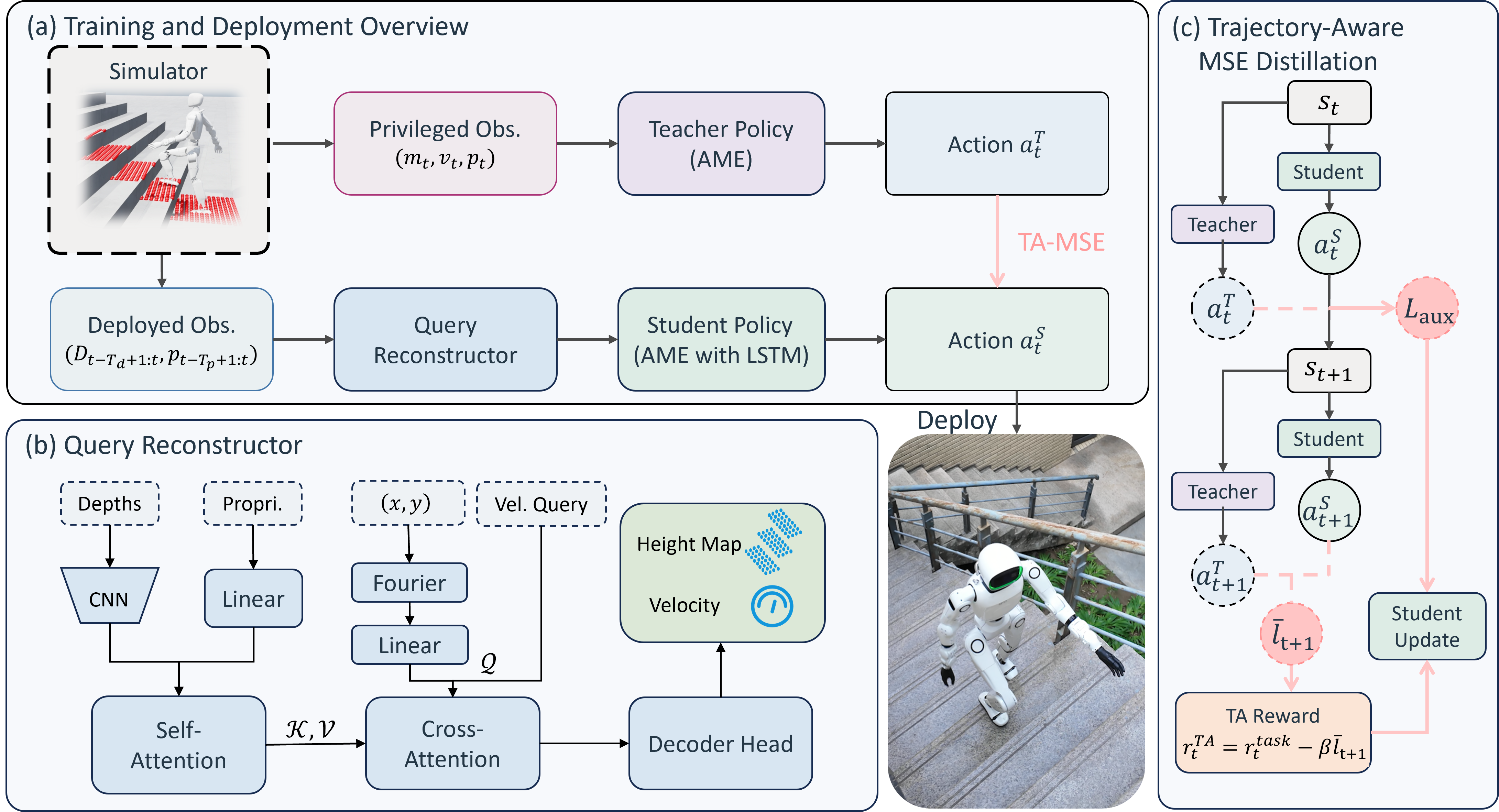}
\caption{\textbf{SOLO overview.}  During training, the privileged teacher
provides action targets on student-visited states.  QR recovers the local
    terrain map and base velocity from depth and proprioceptive histories, and
    \TAMSE{} distills the teacher into the student.  Deployment retains only QR
    and the student policy for zero-shot deployment from onboard sensing during all trials.}
\label{fig:framework}
\end{figure*}

Figure~\ref{fig:framework} summarizes the information flow.  At run time,
depth and proprioceptive histories are encoded into a shared sensor memory,
QR reconstructs $(\hat m_t,\hat v_t)$, and the student maps the reconstructed
state to an action.  The teacher, critic, and privileged observations remain
training-only and are removed completely for onboard policy deployment on
hardware in every hardware trial.

\subsection{Query Reconstructor}
\label{sec:qr}

QR predicts a robot-centric local height map and base velocity from deployable
observations.  Unlike dense decoding from a shared feature, QR gives each cell
a spatial query that retrieves location-specific sensor-history evidence at
that location for every queried map cell in robot coordinates.

\paragraph{Sensor token memory.}
A lightweight ResNet~\citep{he2016resnet} encodes each frame in the depth
history into visual tokens.  A per-frame MLP separately embeds the
proprioceptive history, to which learned temporal position embeddings are
added.  We concatenate both token sets and process them with a Transformer
encoder~\citep{vaswani2017attention}, producing the sensor memory $M_t$.
This shared memory interprets terrain evidence with the robot's motion and
contact-phase context during partially observed locomotion over changing terrain
throughout each rollout with changing viewpoints and contact states.

\paragraph{Cell-query decoding.}
For a map with $N=G_xG_y$ cells, we Fourier-encode the planar coordinate of
each cell~\citep{tancik2020fourier} and project it into a query $q_i$.  The
$N$ cell queries attend to $M_t$ through stacked multi-head
cross-attention~\citep{carion2020detr}; a height head then maps the decoded
query $q_i'$ to the corresponding height $\hat h_i$.  A separate learned
velocity query attends to the same memory and is decoded into the 3-DoF
base-velocity estimate $\hat v_t$.  Map and velocity predictions therefore
share sensor evidence without forcing all cells through a shared dense
bottleneck that can blur localized terrain evidence near sharp foothold
boundaries during foothold selection on sparse terrain transitions online.

\paragraph{Reconstruction objective and on-policy update.}
The simulator supplies the privileged height map and base velocity only as
labels.  We optimize the supervised objective shown below:
\begin{equation}
\mathcal{L}_\text{rec}
= \lambda_h \cdot \tfrac{1}{N}\sum_{i=1}^{N}|\hat h_i-h_i^\text{GT}|
+ \lambda_v \cdot \|\hat v_t-v_t^\text{GT}\|_1.
\label{eq:rec-loss}
\end{equation}
In Stage~III, QR is updated from the QR-conditioned student's on-policy
buffer.  Policy updates consume detached QR predictions, while reconstruction
updates optimize only QR through Eq.~\eqref{eq:rec-loss}.
App.~\ref{app:arch} reports the complete deployed architecture, including its
sensor-history dimensions.

\subsection{Trajectory-Aware MSE Distillation}
\label{sec:clmse}

MSE+PPO combines PPO with current-state action matching.  \TAMSE{} retains
this supervision but penalizes disagreement at the state reached by the
student's action, so the return captures how each decision changes later
teacher matching across the sampled rollout throughout training end to end.

\paragraph{Current-state teacher matching.}
On a student rollout, the standard auxiliary objective is
\begin{equation}
L_\text{aux}(\theta)
= \mathbb{E}_{t}\!\left[
\left\|\mu_\theta(h_t^S)-\mu_T(o_t^T)\right\|_2^2
\right].
\label{eq:clmse-aux}
\end{equation}
It matches the student to the teacher at each visited state, but treats samples
independently without encoding how current actions alter later visited states
during policy rollouts at each policy update.

\paragraph{Next-state disagreement reward.}
After the system moves from $s_t$ to $s_{t+1}$, we
evaluate both policies at that student-visited next state and store the
resulting disagreement directly for PPO:
\begin{equation}
\bar\ell_{t+1}
= \operatorname{sg}\!\left(
\left\|\mu_\theta(h_{t+1}^S)-\mu_T(o_{t+1}^T)\right\|_2^2
\right),
\end{equation}
where $\operatorname{sg}(\cdot)$ stores the disagreement on the preceding
transition producing that next state directly:
\begin{equation}
r_t^\text{TA}=r_t^\text{task}-\beta\,\bar\ell_{t+1}.
\label{eq:clmse-reward}
\end{equation}
This temporal alignment penalizes actions that induce states harder for the
student to match.  Through Generalized Advantage Estimation
(GAE)~\citep{schulman2016gae}, later disagreement penalties influence
preceding actions without another critic, dynamics model, or teacher rollout
during policy optimization over the complete sampled trajectory at every update
step over the complete rollout horizon.

The trajectory-weighted imitation component of the sampled return
for action $a_t$ is
\begin{equation}
G_t^{\mathrm{imit}}
=-\beta\sum_{k=t+1}^{T_R}\gamma^{k-t-1}
\operatorname{sg}[\bar\ell_k],
\label{eq:ta-return}
\end{equation}
where $T_R$ is the terminal or rollout boundary.  ``One-step'' denotes the
attachment point, while ``trajectory-aware'' describes how the return and GAE
propagate later disagreement to earlier actions over the full sampled horizon
used for every subsequent policy update during training.

\paragraph{Unified policy objective.}
The student actor maximizes the combined policy objective below:
\begin{equation}
\max_\theta\;J_\text{PPO}(\theta;r^\text{TA})
-\lambda L_\text{aux}(\theta).
\label{eq:clmse-objective}
\end{equation}
This objective recovers the controller baselines in
\S\ref{sec:exp-distill}: PPO uses $\lambda=0,\beta=0$; MSE+PPO uses
$\lambda>0,\beta=0$; and \TAMSE{} uses $\lambda>0,\beta>0$.  All teacher
labels are queried along the student's rollouts and are already available
in standard teacher--student training pipelines used in practice.

\subsection{Training Procedure and Deployment}
\label{sec:overview}

Training proceeds in three stages.  \textbf{Stage 1} trains the privileged
teacher with AMP-regularized PPO; neither the student nor QR is updated.
\textbf{Stage 2} transfers only shape-compatible \AME{} parameters to the
student; its action head and LSTM bypass are initialized randomly.  The student
is distilled with \TAMSE{} while retaining privileged terrain input, and QR
remains disconnected.  \textbf{Stage 3} replaces the
privileged terrain state with QR estimates, jointly fine-tunes the student,
and updates QR on the student's on-policy buffer as described above during
final on-policy deployment fine-tuning.

Training uses massively parallel Isaac Lab rollouts~\citep{mittal2025isaaclab,
makoviychuk2021isaacgym}, PPO~\citep{schulman2017ppo}, and standard dynamics
and observation randomization for sim-to-real transfer~\citep{tobin2017domainrand,
peng2018dynamicsrand,tan2018simtoreal}.  Deployment removes the teacher,
critic, AMP discriminator, and privileged labels; QR and the student run at
50~Hz from only chest-mounted depth and proprioception.  Apps.~\ref{app:arch}--
\ref{app:platform-hw} list the network, optimization, and hardware details
reported in all experiments and hardware evaluations under every reported protocol.

%===============================================================================
\section{Experiments}
\label{sec:exp}

We evaluate SOLO from system-level deployment to component-level evidence.
Zero-shot hardware trials establish stable long-horizon locomotion in the
wild (\S\ref{sec:real}).  Reconstruction comparisons isolate QR's gains over
DPL and START (\S\ref{sec:exp-recon}); curriculum results test whether
\TAMSE{} reaches higher terrain levels faster than PPO and MSE+PPO
(\S\ref{sec:exp-distill}); and highest-difficulty success and foothold-quality
diagnostics test whether these gains translate into long-horizon locomotion
(\S\ref{sec:exp-e2e}, Figure~\ref{fig:recon-and-foothold-quality}) under the matched
experimental protocols used consistently for all comparisons.

\subsection{Real-Robot Deployment}
\label{sec:real}

\paragraph{Deployment protocol.}  SOLO is command-conditioned: a human operator
provides only the planar joystick velocity command $c_t$.  The student maps it,
four-frame D455 depth, and ten-frame proprioceptive/control history to 25 joint
targets.  We deploy zero-shot on Omni without external state or mapping on an
outdoor long-horizon route and an indoor mixed-terrain course;
App.~\ref{app:platform-hw} reports the hardware and runtime details for all trials
on the same physical Omni platform.

\begin{wraptable}[10]{r}{0.35\textwidth}
\vspace{-1.5em}
\centering
\caption{\textbf{Real-world isolated-terrain success} over ten trials per
terrain on the hardware.}
\label{tab:real-success}
\vspace{0.5em}
\scriptsize
\renewcommand{\arraystretch}{0.96}
\begin{tabular*}{\linewidth}{@{\extracolsep{\fill}}cc@{}}
\toprule
Terrain & Success ($\uparrow$) \\
\midrule
Ascending stairs         & $10/10$ \\
Descending stairs        & $10/10$ \\
Slope ($20^\circ$)       & $10/10$ \\
High platform ($40$\,cm) & $10/10$ \\
Gap ($50$\,cm)           & $10/10$ \\
Movable obstacle         & $10/10$ \\
Stepping stones          & $9/10$  \\
\midrule
Overall                  & $69/70$ \\
\bottomrule
\end{tabular*}
\vspace{-0.3em}
\end{wraptable}

\paragraph{Outdoor long-horizon deployment.} The outdoor experiment evaluates real-world locomotion along a natural route containing stairs, slopes, grass transitions, and uneven ground. Figure~\ref{fig:stable_outdoor_perceptive_locomotion} shows a single continuous 1.5~km deployment completed without policy resets or physical assistance.

\begin{figure*}[!t]
\centering
\includegraphics[width=0.88\textwidth]{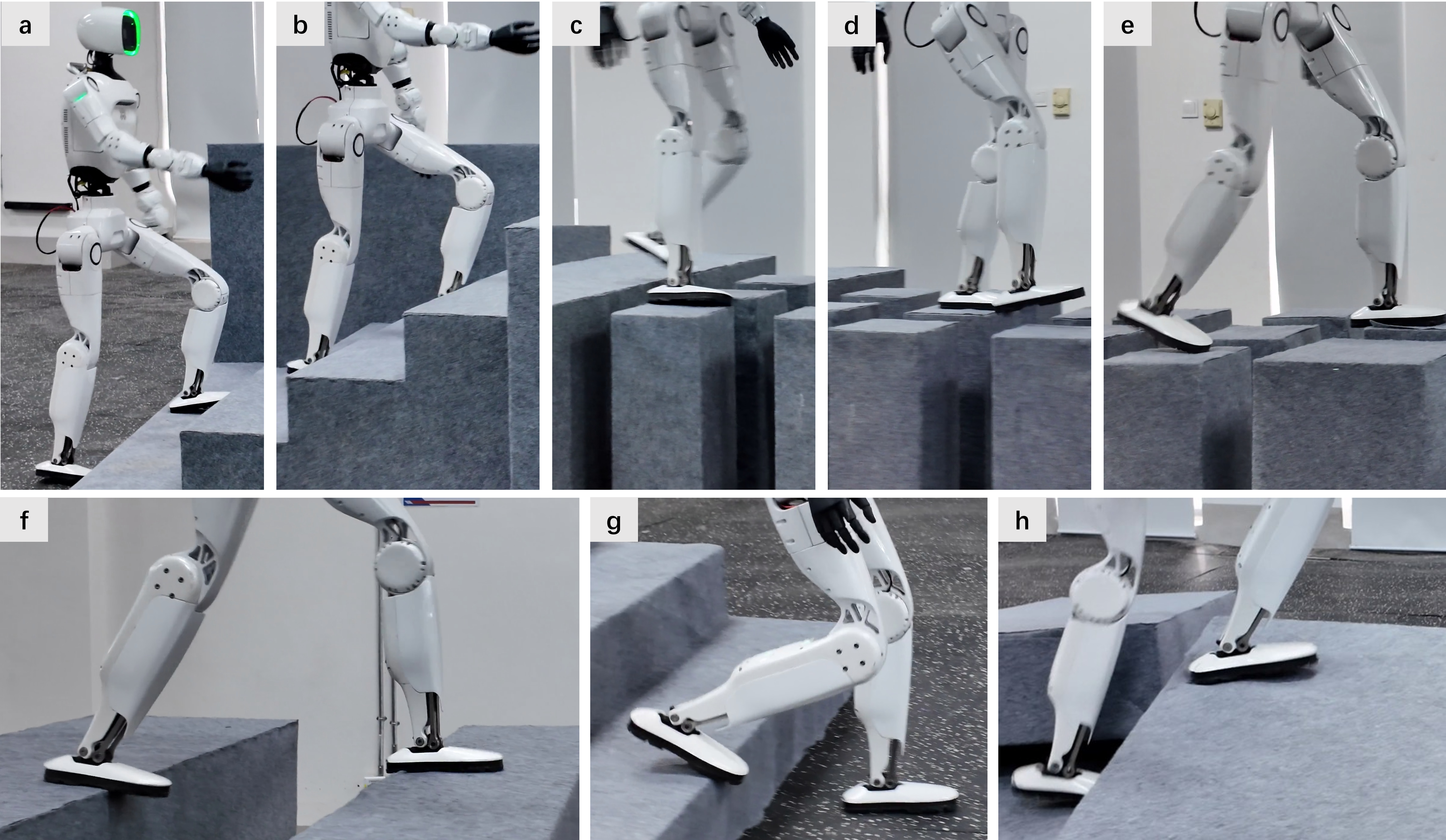}
\caption{\textbf{Continuous indoor mixed-terrain deployment.}  One run
traverses ascending stairs, stepping stones, a gap,
descending stairs, and a movable obstacle without policy resets in one run.}
\label{fig:indoor-mixed-deployment}
\end{figure*}

\paragraph{Indoor mixed-terrain deployment.}  One run concentrates
terrain-interaction challenges: ascending stairs,
stepping stones, a gap, descending stairs and a movable obstacle.
Figure~\ref{fig:indoor-mixed-deployment} shows the uninterrupted traversal;
Appendix~Figure~\ref{fig:indoor-terrain-setup} shows the layout; every obstacle
is negotiated without a reset or physical intervention
by the same SOLO policy throughout the uninterrupted course.

\noindent\textbf{Isolated-terrain reliability.}
We complement the continuous deployments with ten trials on each of seven
isolated real-world terrains, defining success as traversal without a fall,
policy reset, or physical assistance.  Table~\ref{tab:real-success} records
$69/70$ successes, with the only failure on stepping stones, while all other
terrains achieve ten successes without reset or assistance across all hardware trials.

\begin{table*}[!t]
\centering
\caption{\textbf{Reported real-robot perceptive-locomotion context.}
Hardware perception, sparse-terrain capability, and long-horizon hardware
evidence reported by each system under its hardware setup.}
\label{tab:reported-system-context}
\small
\renewcommand{\arraystretch}{1.25}
\setlength{\tabcolsep}{4pt}
\begin{tabular*}{\textwidth}{@{\extracolsep{\fill}}llll@{}}
\toprule
Method & Hardware perception & Sparse terrain & Long-horizon hardware evidence \\
\midrule
AME (GR-1)~\citep{he2025ame1} & Qualisys + mesh & stones/beams & Not reported \\
CReF~\citep{hao2026cref} & 1 depth camera & stairs/gaps & Not reported \\
RPL~\citep{zhang2026rpl} & 2 depth (4 in sim.) & stepping stones & 50\,m course + curved stairs \\
\textbf{SOLO} & \textbf{1 depth camera} & \textbf{stepping stones} &
\textbf{1.5\,km outdoor + $>$100+ stairs} \\
\bottomrule
\end{tabular*}
\end{table*}

\noindent\textbf{Context among reported hardware evaluations.}
Table~\ref{tab:reported-system-context} compares the sensing configuration,
sparse-terrain capability, and long-horizon hardware evidence reported for
AME, CReF, RPL, and SOLO.  AME uses external Qualisys tracking and a mesh;
CReF reports one depth camera and stair/gap traversal; and RPL reports a
50-m course and curved stairs with two hardware depth cameras.  SOLO uses one
onboard depth camera for stepping-stone locomotion, a continuous 1.5-km outdoor
route, and an ascent of more than 100 stairs in one uninterrupted deployment.

% The deployment results above establish the overall capability of the
% combined system.  The remaining experiments isolate where this capability
% comes from by comparing alternative reconstructors and distillation
% objectives in simulation.

\subsection{Reconstruction Accuracy}
\label{sec:exp-recon}

\paragraph{Setup and metric.}  We isolate the reconstructor while fixing the
teacher, task, curriculum, training budget and distillation objective.  Each
reconstructor is distilled with its own student under the shared
highest-difficulty protocol (App.~\ref{app:stress-env}), using height-map L1
as the metric (App.~\ref{app:terrain-recon-eval}) to quantify per-cell
terrain-state reconstruction accuracy under the same fully matched protocol throughout.

\paragraph{Baselines.}
We compare the START-style recurrent dense decoder~\citep{yu2026start},
DPL-style cross-attention reconstructor~\citep{sun2025dpl}, and QR (ours) under
the shared setup and evaluation protocol in all runs.

\begin{table}[!htbp]
\centering
\caption{\textbf{On-policy height-map reconstruction error.}  Per-cell L1
in cm at curriculum difficulty $0.99$ ($\downarrow$).  Down/Up-25/35 denote
stair direction and tread width for every method and terrain.}
\label{tab:recon}
{\scriptsize
\setlength{\tabcolsep}{3pt}
\renewcommand{\arraystretch}{1.08}
\begin{tabular}{@{}lccccccccc@{}}
\toprule
Method & Gap & Random Grid & Down-25 & Down-35 & Up-25 & Up-35
& Step Stones & High Plane & Avg. \\
\midrule
START~\citep{yu2026start}
& 10.11 & 5.18 & 8.35 & 5.32 & 8.10 & 5.54 & 14.76 & 3.31 & 7.59 \\
DPL~\citep{sun2025dpl}
& 10.62 & 8.09 & 11.74 & 7.47 & 10.68 & 7.57 & 12.75 & 5.16 & 9.26 \\
QR (ours)
& \textbf{2.27} & \textbf{2.43} & \textbf{2.03} & \textbf{1.47}
& \textbf{1.89} & \textbf{1.31} & \textbf{5.92} & \textbf{0.97}
& \textbf{2.29} \\
\bottomrule
\end{tabular}
}
\end{table}

\begin{figure}[!htbp]
\centering
\includegraphics[width=\linewidth]{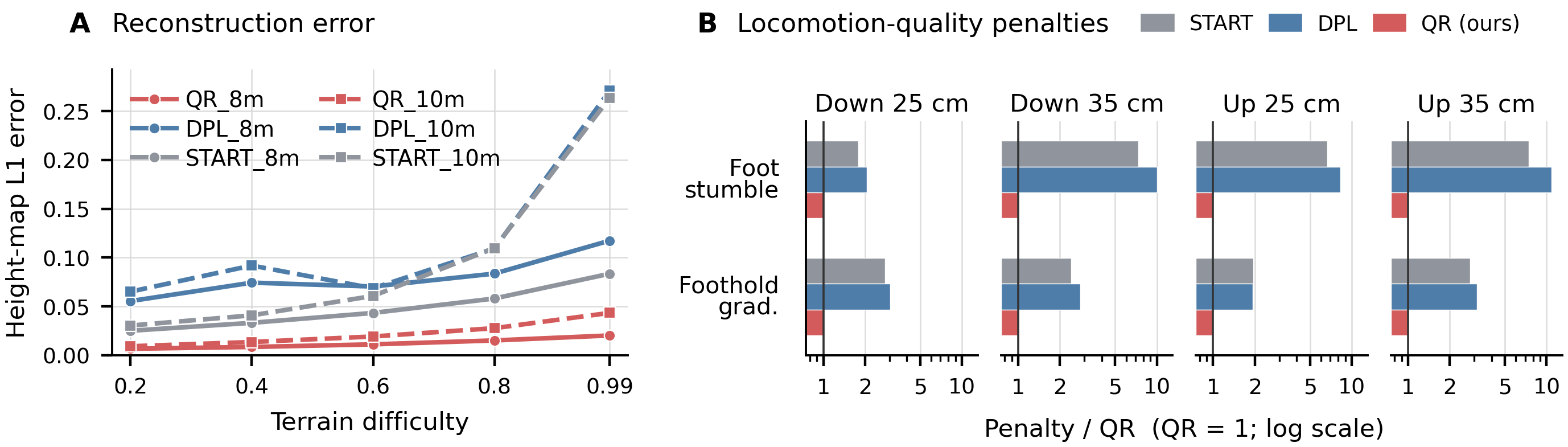}
\caption{\textbf{Reconstruction fidelity and downstream foothold quality.}
All reconstructors use the same teacher, task, curriculum, training budget
and \TAMSE{} objective and are trained on $8$\,m scenes.  (A) Descending-stair
height-map L1 error across curriculum difficulty, testing robustness to a
change in terrain extent; solid circles and dashed squares denote evaluation
on $8$\,m and unseen $10$\,m scenes.  (B) Foot
Stumble and Foothold-Grad penalty magnitudes on highest-difficulty simulated
stairs, normalized by QR for each terrain and term (QR${}=1$; log scale;
lower is better throughout for all four stair settings under the matched
evaluation protocol used throughout).}
\label{fig:recon-and-foothold-quality}
\end{figure}

\paragraph{Results.}  QR reduces average height-map L1 error from
$7.59$/$9.26$\,cm (START/DPL) to $2.29$\,cm, a $3.3{-}4.0\times$ reduction.
The largest absolute reduction occurs on stepping stones, where local
height discontinuities make per-cell fidelity especially important.
Under the matched offline protocol in Appendix~Table~\ref{tab:offline-recon},
QR records $2.98\pm0.07$\,cm overall L1 and $0.386\pm0.008$ Edge F1@1,
compared with $4.32$--$4.36$\,cm and $0.214$--$0.242$ for START and DPL.
Figure~\ref{fig:recon-and-foothold-quality}A tests robustness to a change in
terrain extent on descending stairs.  All reconstructors are trained on
$8$\,m scenes; solid and dashed curves evaluate matched $8$\,m and unseen
$10$\,m scenes.  At difficulty $0.99$, QR changes from $0.020$ to $0.043$\,m,
whereas DPL rises from $0.117$ to $0.271$\,m and START from $0.084$ to
$0.264$\,m.  On the unseen $10$\,m scenes, DPL and START incur $6.2\times$
and $6.1\times$ QR's error, respectively.  The widening cross-size gap shows
that QR is more robust to terrain-extent shift; the baselines overfit the
$8$\,m training geometry and their errors increase sharply at $10$\,m under
the same curriculum difficulty.

Appendix~Figure~\ref{fig:height-map-qualitative} complements the aggregate errors
with highest-difficulty reconstructions.  QR preserves discrete footholds
and sharp height transitions more faithfully, with the clearest difference
on stepping stones across the complete terrain set evaluated at the same difficulty throughout.

\subsection{Distillation Curriculum Progress}
\label{sec:exp-distill}

\paragraph{Setup and metric.}  In the second distillation stage, before QR
is introduced, we use the same teacher, privileged map input, terrain
curriculum, initialization, and training budget to compare PPO only,
MSE+PPO/D-PPO~\citep{
zhang2025dppo, cheng2024parkour, zhang2026ame2, rudin2025parkour}, and
\TAMSE.  The metric is the mean curriculum level reached during training;
higher levels correspond to harder terrain configurations.  Solving the
maximum level triggers level-randomized reassignment, so policies that
consistently solve the hardest setting oscillate near a mean ceiling of $5.5$
under the shared training protocol used for comparison.

\FloatBarrier
\begin{wrapfigure}[13]{r}{0.44\textwidth}
\vspace{-0.8em}
\centering
\includegraphics[width=\linewidth]{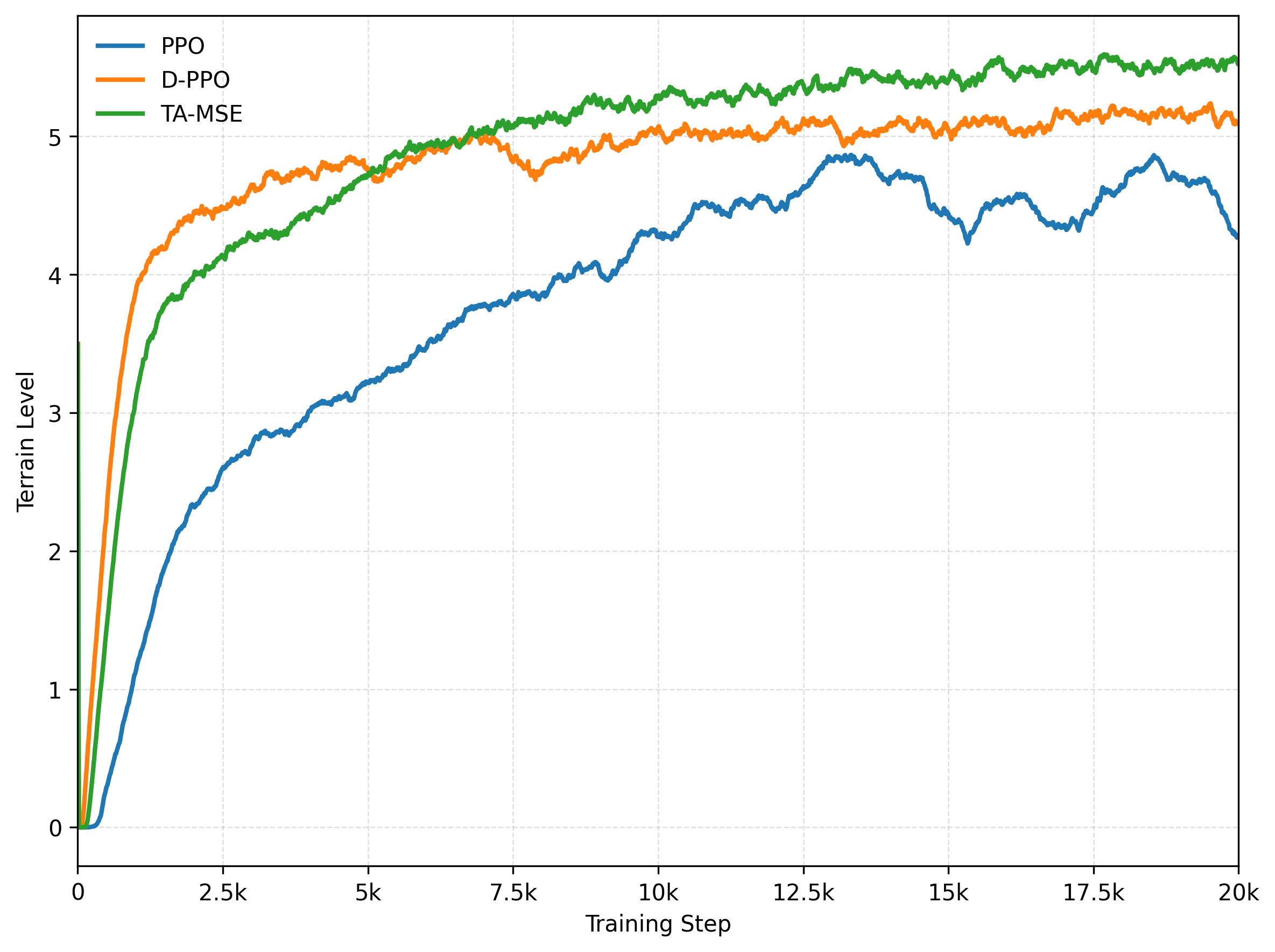}
\caption{Curriculum progress with a matched teacher and privileged map input.}
\label{fig:distill-curve}
\vspace{-0.4em}
\end{wrapfigure}

\paragraph{Results.}
Figure~\ref{fig:distill-curve} shows that \TAMSE{} reaches the highest
curriculum level and keeps improving after the other objectives plateau.
\TAMSE{} reaches the curriculum ceiling at about $5.5$, while MSE+PPO
plateaus around $5.1$ and PPO remains below $5.0$ with larger oscillations.
All three objectives advance rapidly at lower levels, but their trajectories
separate as the curriculum approaches its hardest configurations.  \TAMSE{}
continues to gain after both baselines stabilize, so return-based future
disagreement supplies its strongest additional credit signal throughout all
training runs at the hardest levels evaluated.

\subsection{Locomotion on Highest-Difficulty Terrain}
\label{sec:exp-e2e}

\paragraph{Setup and metrics.}
Under the shared highest-difficulty protocol in App.~\ref{app:stress-env},
we compare reconstructors with \TAMSE{} fixed.  The main metric is
per-terrain success rate.  Foot Stumble measures foot--terrain impacts where
horizontal contact force dominates vertical support, while Foothold-Grad
penalizes feet placed across local height discontinuities
(Figure~\ref{fig:recon-and-foothold-quality}B).  Both provide foothold-quality
proxies beyond binary success across all evaluated terrain configurations with
identical commands, training budgets, and termination criteria for each matched
evaluation run in the shared simulator.

\paragraph{Baselines.}
We compare \TAMSE{} with (C1) START~\citep{yu2026start}, (C2)
DPL~\citep{sun2025dpl}, and (C3) QR (ours) under the shared protocol, with all
other experimental controls fixed throughout the evaluation.

\begin{figure}[!htbp]
\centering
\includegraphics[width=\linewidth]{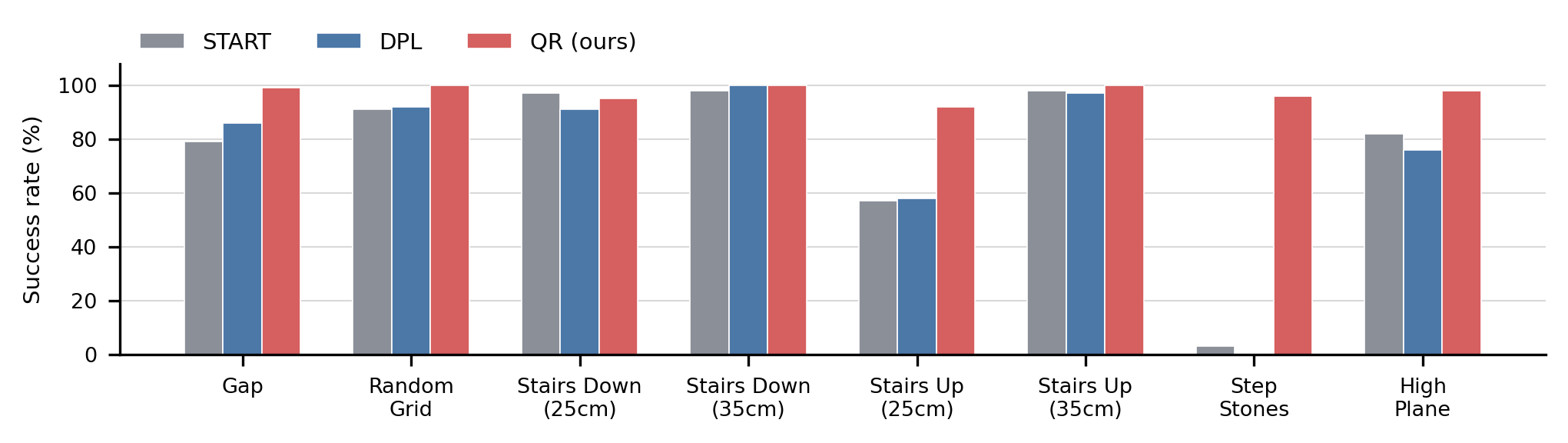}
\caption{Highest-difficulty success rate for $100$ rollouts per terrain at
curriculum difficulty $0.99$.  Stair labels denote tread width; Up-35 is
the ascending-stair configuration with $0.35$\,m tread width, and High
Plane contains two $0.35$\,m-high platforms in this evaluation for every method shown.}
\label{fig:highest-difficulty-success}
\end{figure}

\paragraph{Results.}
As shown in Figure~\ref{fig:highest-difficulty-success}, QR improves mean
success from $75.0$--$75.6\%$ to $97.5\%$ and raises worst-case terrain
success from $0$--$3\%$ to $92\%$.  The gain is largest on sparse
footholds: on stepping stones QR reaches $96\%$ while START and DPL reach
$3\%$ and $0\%$.  Figure~\ref{fig:recon-and-foothold-quality}B further shows
that QR has the lowest Foot Stumble and Foothold-Grad penalty magnitudes in
all four stair settings.  Across the two terms, START and DPL incur
$1.78$--$7.40\times$ and $1.94$--$10.83\times$ QR's penalties.  The largest
Foot Stumble gaps occur on the $35$\,cm stairs: START reaches
$7.37$/$7.40\times$ and DPL reaches $10.08$/$10.83\times$ QR on descending/
ascending stairs across every matched terrain stress test.

\paragraph{Analysis.}
Across the comparisons, QR combines lower reconstruction error
(Table~\ref{tab:recon} and
Figure~\ref{fig:recon-and-foothold-quality}A),
lower foothold-quality penalties
(Figure~\ref{fig:recon-and-foothold-quality}B), and higher
highest-difficulty success
(Figure~\ref{fig:highest-difficulty-success}).
The relationship is strongest on sparse footholds and stairs, where safe
support regions lie beside sharp height discontinuities.  The qualitative
maps in Appendix~Figure~\ref{fig:height-map-qualitative} show the same
pattern: retaining local geometry improves foothold quality under the same
shared evaluation protocol.

\FloatBarrier

\section{Conclusion}
\label{sec:conclusion}

SOLO combines per-cell terrain reconstruction and trajectory-aware
distillation for stable long-horizon perceptive humanoid locomotion.  QR
reduces highest-difficulty height-map L1 error by $3.3$--$4.0\times$ relative
to START and DPL.  \TAMSE{} routes next-state disagreement through the PPO
return and reaches the curriculum ceiling after PPO and MSE+PPO plateau.  The
student is distilled with privileged terrain input in Stage~II; Stage~III
introduces QR and supervises it on states visited by the current QR-conditioned
student.  With \TAMSE{} fixed, QR raises mean stress-test success from
$75.0$--$75.6\%$ to $97.5\%$ and stepping-stone success from $0$--$3\%$ to
$96\%$.  The same D455--proprioception--QR--student stack supports the 1.5-km
route, indoor mixed-terrain course, and $>$10-floor ascent; six isolated
terrains achieve $10/10$ success and stepping stones achieve $9/10$.  All
hardware trials retain the same 50-Hz, 25-joint action interface on Omni
hardware.

\section{Limitations}
\label{sec:limit}

SOLO is command-conditioned: a human supplies planar velocity commands, while
the learned policy handles terrain perception and whole-body control.  Its
front-facing depth camera leaves a rear blind spot during backward locomotion
on changing terrain.  Longer-lived memory, as in LF2WB~\citep{luo2026lf2wb},
retains foothold geometry beyond the camera view.  The 2.5D height map
represents one ground surface per cell, and transparent or reflective surfaces
can corrupt depth.  Complementary sensing and volumetric terrain
representations extend this pipeline to omnidirectional deployment over these
challenging settings while retaining the command-conditioned policy interface.

% %===============================================================================
% \clearpage
% \acknowledgments{\PH{To be added in the camera-ready version.}}

%===============================================================================
\clearpage
\bibliography{paper}

@inproceedings{czarnecki2019distilling,
  title     = {Distilling Policy Distillation},
  author    = {Czarnecki, Wojciech M. and Pascanu, Razvan and Osindero, Simon
               and Jayakumar, Siddhant and Swirszcz, Grzegorz and Jaderberg, Max},
  booktitle = {Proceedings of the Twenty-Second International Conference on
               Artificial Intelligence and Statistics},
  pages     = {1331--1340},
  year      = {2019},
  volume    = {89},
  series    = {Proceedings of Machine Learning Research},
  publisher = {PMLR},
  url       = {https://proceedings.mlr.press/v89/czarnecki19a.html}
}

@inproceedings{rusu2015policy,
  title     = {Policy Distillation},
  author    = {Rusu, Andrei A. and Colmenarejo, Sergio Gomez and
               G{\"u}l{\c{c}}ehre, {\c{C}}aglar and Desjardins, Guillaume
               and Kirkpatrick, James and Pascanu, Razvan and Mnih, Volodymyr
               and Kavukcuoglu, Koray and Hadsell, Raia},
  booktitle = {International Conference on Learning Representations},
  year      = {2016},
  url       = {https://mlanthology.org/iclr/2016/rusu2016iclr-policy/}
}

@article{schmitt2018kickstart,
  title         = {Kickstarting Deep Reinforcement Learning},
  author        = {Schmitt, Simon and Hudson, Jonathan J. and Zidek, Augustin
                   and Osindero, Simon and Doersch, Carl and Czarnecki, Wojciech M.
                   and Leibo, Joel Z. and Kuttler, Heinrich and Zisserman, Andrew
                   and Simonyan, Karen and Eslami, S. M. Ali},
  journal       = {arXiv preprint arXiv:1803.03835},
  year          = {2018},
  eprint        = {1803.03835},
  archiveprefix = {arXiv},
  primaryclass  = {cs.LG},
  url           = {https://arxiv.org/abs/1803.03835}
}

@inproceedings{ross2011dagger,
  title     = {A Reduction of Imitation Learning and Structured Prediction to
               No-Regret Online Learning},
  author    = {Ross, Stephane and Gordon, Geoffrey and Bagnell, Drew},
  booktitle = {Proceedings of the Fourteenth International Conference on
               Artificial Intelligence and Statistics},
  pages     = {627--635},
  year      = {2011},
  volume    = {15},
  series    = {Proceedings of Machine Learning Research},
  publisher = {PMLR},
  url       = {https://proceedings.mlr.press/v15/ross11a.html}
}

@inproceedings{hinton2015kd,
  title     = {Distilling the Knowledge in a Neural Network},
  author    = {Hinton, Geoffrey and Vinyals, Oriol and Dean, Jeff},
  booktitle = {NeurIPS Deep Learning and Representation Learning Workshop},
  year      = {2015},
  url       = {https://research.google/pubs/distilling-the-knowledge-in-a-neural-network/}
}

@inproceedings{zhang2025dppo,
  title     = {Distillation-{PPO}: A Novel Two-Stage Reinforcement Learning
               Framework for Humanoid Robot Perceptive Locomotion},
  author    = {Zhang, Qiang and Han, Gang and Sun, Jingkai and Zhao, Wen
               and Sun, Chenghao and Cao, Jiahang and Wang, Jiaxu and Guo, Yijie
               and Xu, Renjing},
  booktitle = {2025 IEEE/RSJ International Conference on Intelligent Robots and
               Systems (IROS)},
  pages     = {2916--2922},
  year      = {2025},
  month     = {October},
  publisher = {IEEE},
  doi       = {10.1109/IROS60139.2025.11245929},
  url       = {https://ieeexplore.ieee.org/document/11245929}
}

@article{he2025ame1,
  title   = {Attention-Based Map Encoding for Learning Generalized Legged Locomotion},
  author  = {He, Junzhe and Zhang, Chong and Jenelten, Fabian and Grandia, Ruben
             and B{\"a}cher, Moritz and Hutter, Marco},
  journal = {Science Robotics},
  volume  = {10},
  number  = {105},
  pages   = {eadv3604},
  year    = {2025},
  month   = {August},
  doi     = {10.1126/scirobotics.adv3604},
  url     = {https://www.science.org/doi/10.1126/scirobotics.adv3604}
}

@article{zhang2026ame2,
  title         = {{AME-2}: Agile and Generalized Legged Locomotion via
                   Attention-Based Neural Map Encoding},
  author        = {Zhang, Chong and Klemm, Victor and Yang, Fan and Hutter, Marco},
  journal       = {arXiv preprint arXiv:2601.08485},
  year          = {2026},
  eprint        = {2601.08485},
  archiveprefix = {arXiv},
  primaryclass  = {cs.RO},
  url           = {https://arxiv.org/abs/2601.08485}
}

@article{miki2022wild,
  title   = {Learning Robust Perceptive Locomotion for Quadrupedal Robots in the Wild},
  author  = {Miki, Takahiro and Lee, Joonho and Hwangbo, Jemin and Wellhausen, Lorenz
             and Koltun, Vladlen and Hutter, Marco},
  journal = {Science Robotics},
  volume  = {7},
  number  = {62},
  pages   = {eabk2822},
  year    = {2022},
  month   = {January},
  doi     = {10.1126/scirobotics.abk2822},
  url     = {https://www.science.org/doi/10.1126/scirobotics.abk2822}
}

@article{lee2020quadruped,
  title   = {Learning Quadrupedal Locomotion over Challenging Terrain},
  author  = {Lee, Joonho and Hwangbo, Jemin and Wellhausen, Lorenz and Koltun, Vladlen
             and Hutter, Marco},
  journal = {Science Robotics},
  volume  = {5},
  number  = {47},
  pages   = {eabc5986},
  year    = {2020},
  month   = {October},
  doi     = {10.1126/scirobotics.abc5986},
  url     = {https://www.science.org/doi/10.1126/scirobotics.abc5986}
}

@article{sun2025dpl,
  title         = {{DPL}: Depth-only Perceptive Humanoid Locomotion via Realistic
                   Depth Synthesis and Cross-Attention Terrain Reconstruction},
  author        = {Sun, Jingkai and Han, Gang and Sun, Pihai and Zhao, Wen
                   and Cao, Jiahang and Wang, Jiaxu and Guo, Yijie and Zhang, Qiang},
  journal       = {arXiv preprint arXiv:2510.07152},
  year          = {2025},
  eprint        = {2510.07152},
  archiveprefix = {arXiv},
  primaryclass  = {cs.RO},
  url           = {https://arxiv.org/abs/2510.07152}
}

@article{yu2026start,
  title   = {{START}: Traversing Sparse Footholds With Terrain Reconstruction},
  author  = {Yu, Ruiqi and Wang, Qianshi and Li, Hongyi and Jun, Zheng
             and Wang, Zhicheng and Wu, Jun and Zhu, Qiuguo},
  journal = {IEEE Robotics and Automation Letters},
  volume  = {11},
  number  = {2},
  pages   = {2194--2201},
  year    = {2026},
  month   = {February},
  doi     = {10.1109/LRA.2025.3645649},
  url     = {https://ieeexplore.ieee.org/document/11303867}
}

@article{hoeller2024parkour,
  title   = {{ANYmal} Parkour: Learning Agile Navigation for Quadrupedal Robots},
  author  = {Hoeller, David and Rudin, Nikita and Sako, Dhionis and Hutter, Marco},
  journal = {Science Robotics},
  volume  = {9},
  number  = {88},
  pages   = {eadi7566},
  year    = {2024},
  month   = {March},
  doi     = {10.1126/scirobotics.adi7566},
  url     = {https://www.science.org/doi/10.1126/scirobotics.adi7566}
}

@inproceedings{yang2023nvm,
  title     = {Neural Volumetric Memory for Visual Locomotion Control},
  author    = {Yang, Ruihan and Yang, Ge and Wang, Xiaolong},
  booktitle = {Proceedings of the IEEE/CVF Conference on Computer Vision and
               Pattern Recognition},
  pages     = {1430--1440},
  year      = {2023},
  url       = {https://openaccess.thecvf.com/content/CVPR2023/html/Yang_Neural_Volumetric_Memory_for_Visual_Locomotion_Control_CVPR_2023_paper.html}
}

@inproceedings{wang2025beamdojo,
  title     = {{BeamDojo}: Learning Agile Humanoid Locomotion on Sparse Footholds},
  author    = {Wang, Huayi and Wang, Zirui and Ren, Junli and Ben, Qingwei
               and Huang, Tao and Zhang, Weinan and Pang, Jiangmiao},
  booktitle = {Proceedings of Robotics: Science and Systems},
  year      = {2025},
  month     = {June},
  address   = {Los Angeles, CA, USA},
  doi       = {10.15607/RSS.2025.XXI.068},
  url       = {https://www.roboticsproceedings.org/rss21/p068.html}
}

@article{zhang2026rpl,
  title         = {{RPL}: Learning Robust Humanoid Perceptive Locomotion on
                   Challenging Terrains},
  author        = {Zhang, Yuanhang and Seo, Younggyo and Chen, Juyue and Yuan, Yifu
                   and Sreenath, Koushil and Abbeel, Pieter and Sferrazza, Carmelo
                   and Liu, Karen and Duan, Rocky and Shi, Guanya},
  journal       = {arXiv preprint arXiv:2602.03002},
  year          = {2026},
  eprint        = {2602.03002},
  archiveprefix = {arXiv},
  primaryclass  = {cs.RO},
  url           = {https://arxiv.org/abs/2602.03002}
}

@inproceedings{cheng2024parkour,
  title     = {Extreme Parkour with Legged Robots},
  author    = {Cheng, Xuxin and Shi, Kexin and Agarwal, Ananye and Pathak, Deepak},
  booktitle = {2024 IEEE International Conference on Robotics and Automation},
  pages     = {11443--11450},
  year      = {2024},
  month     = {May},
  doi       = {10.1109/ICRA57147.2024.10610200},
  url       = {https://ieeexplore.ieee.org/document/10610200}
}

@inproceedings{zhuang2025humanoidparkour,
  title     = {Humanoid Parkour Learning},
  author    = {Zhuang, Ziwen and Yao, Shenzhe and Zhao, Hang},
  booktitle = {Proceedings of the 8th Conference on Robot Learning},
  pages     = {1975--1991},
  year      = {2025},
  volume    = {270},
  series    = {Proceedings of Machine Learning Research},
  publisher = {PMLR},
  url       = {https://proceedings.mlr.press/v270/zhuang25a.html}
}

@article{rudin2025parkour,
  title         = {Parkour in the Wild: Learning a General and Extensible Agile
                   Locomotion Policy Using Multi-Expert Distillation and
                   {RL} Fine-Tuning},
  author        = {Rudin, Nikita and He, Junzhe and Aurand, Joshua and Hutter, Marco},
  journal       = {arXiv preprint arXiv:2505.11164},
  year          = {2025},
  eprint        = {2505.11164},
  archiveprefix = {arXiv},
  primaryclass  = {cs.RO},
  url           = {https://arxiv.org/abs/2505.11164}
}

@incollection{carion2020detr,
  title     = {End-to-End Object Detection with Transformers},
  author    = {Carion, Nicolas and Massa, Francisco and Synnaeve, Gabriel
               and Usunier, Nicolas and Kirillov, Alexander and Zagoruyko, Sergey},
  booktitle = {Computer Vision -- ECCV 2020},
  pages     = {213--229},
  year      = {2020},
  publisher = {Springer International Publishing},
  doi       = {10.1007/978-3-030-58452-8_13},
  url       = {https://doi.org/10.1007/978-3-030-58452-8_13}
}

@article{hochreiter1997lstm,
  title   = {Long Short-Term Memory},
  author  = {Hochreiter, Sepp and Schmidhuber, J{\"u}rgen},
  journal = {Neural Computation},
  volume  = {9},
  number  = {8},
  pages   = {1735--1780},
  year    = {1997},
  month   = {November},
  doi     = {10.1162/neco.1997.9.8.1735},
  url     = {https://doi.org/10.1162/neco.1997.9.8.1735}
}

@article{li2024humanoidlearningfundamental,
  title   = {Reinforcement Learning for Versatile, Dynamic, and Robust Bipedal
             Locomotion Control},
  author  = {Li, Zhongyu and Peng, Xue Bin and Abbeel, Pieter and Levine, Sergey
             and Berseth, Glen and Sreenath, Koushil},
  journal = {The International Journal of Robotics Research},
  volume  = {44},
  number  = {5},
  pages   = {840--888},
  year    = {2025},
  month   = {April},
  doi     = {10.1177/02783649241285161},
  url     = {https://doi.org/10.1177/02783649241285161}
}

@article{mittal2025isaaclab,
  title   = {Isaac Lab: A {GPU}-Accelerated Simulation Framework for
             Multi-Modal Robot Learning},
  author  = {Mayank Mittal and Pascal Roth and James Tigue and Antoine Richard
             and Octi Zhang and Peter Du and Antonio Serrano-Mu{\~n}oz
             and Xinjie Yao and Ren{\'e} Zurbr{\"u}gg and Nikita Rudin
             and Lukasz Wawrzyniak and Milad Rakhsha and Alain Denzler
             and Eric Heiden and Ales Borovicka and Ossama Ahmed
             and Iretiayo Akinola and Abrar Anwar and Mark T. Carlson
             and Ji Yuan Feng and Animesh Garg and Renato Gasoto
             and Lionel Gulich and Yijie Guo and M. Gussert
             and Alex Hansen and Mihir Kulkarni and Chenran Li
             and Wei Liu and Viktor Makoviychuk and Grzegorz Malczyk
             and Hammad Mazhar and Masoud Moghani
             and Adithyavairavan Murali and Michael Noseworthy
             and Alexander Poddubny and Nathan Ratliff and Welf Rehberg
             and Clemens Schwarke and Ritvik Singh and James Latham Smith
             and Bingjie Tang and Ruchik Thaker and Matthew Trepte
             and Karl Van Wyk and Fangzhou Yu and Alex Millane
             and Vikram Ramasamy and Remo Steiner and Sangeeta Subramanian
             and Clemens Volk and CY Chen and Neel Jawale
             and Ashwin Varghese Kuruttukulam and Michael A. Lin
             and Ajay Mandlekar and Karsten Patzwaldt and John Welsh
             and Huihua Zhao and Fatima Anes and Jean-Francois Lafleche
             and Nicolas Mo{\"e}nne-Loccoz and Soowan Park
             and Rob Stepinski and Dirk Van Gelder and Chris Amevor
             and Jan Carius and Jumyung Chang and Anka He Chen
             and Pablo de Heras Ciechomski and Gilles Daviet
             and Mohammad Mohajerani and Julia von Muralt
             and Viktor Reutskyy and Michael Sauter and Simon Schirm
             and Eric L. Shi and Pierre Terdiman and Kenny Vilella
             and Tobias Widmer and Gordon Yeoman and Tiffany Chen
             and Sergey Grizan and Cathy Li and Lotus Li and Connor Smith
             and Rafael Wiltz and Kostas Alexis and Yan Chang and David Chu
             and {Linxi "Jim" Fan} and Farbod Farshidian and Ankur Handa
             and Spencer Huang and Marco Hutter and Yashraj Narang
             and Soha Pouya and Shiwei Sheng and Yuke Zhu and Miles Macklin
             and Adam Moravanszky and Philipp Reist and Yunrong Guo
             and David Hoeller and Gavriel State},
  journal = {arXiv preprint arXiv:2511.04831},
  year    = {2025},
  url     = {https://arxiv.org/abs/2511.04831}
}

@article{makoviychuk2021isaacgym,
  title   = {Isaac Gym: High Performance {GPU}-Based Physics Simulation for
             Robot Learning},
  author  = {Makoviychuk, Viktor and Wawrzyniak, Lukasz and Guo, Yunrong and
             Lu, Michelle and Storey, Kier and Macklin, Miles and Hoeller,
             David and Rudin, Nikita and Allshire, Arthur and Handa, Ankur
             and State, Gavriel},
  journal = {arXiv preprint arXiv:2108.10470},
  year    = {2021},
  url     = {https://arxiv.org/abs/2108.10470}
}

@article{peng2021amp,
  title   = {{AMP}: Adversarial Motion Priors for Stylized Physics-Based
             Character Control},
  author  = {Peng, Xue Bin and Ma, Ze and Abbeel, Pieter and Levine, Sergey
             and Kanazawa, Angjoo},
  journal = {ACM Transactions on Graphics},
  volume  = {40},
  number  = {4},
  pages   = {1--20},
  year    = {2021},
  month   = {July},
  doi     = {10.1145/3450626.3459670},
  url     = {https://doi.org/10.1145/3450626.3459670}
}

@misc{schulman2017ppo,
  title         = {Proximal Policy Optimization Algorithms},
  author        = {John Schulman and Filip Wolski and Prafulla Dhariwal
                   and Alec Radford and Oleg Klimov},
  year          = {2017},
  eprint        = {1707.06347},
  archiveprefix = {arXiv},
  primaryclass  = {cs.LG},
  url           = {https://arxiv.org/abs/1707.06347}
}

@inproceedings{schulman2016gae,
  author       = {John Schulman and
                  Philipp Moritz and
                  Sergey Levine and
                  Michael I. Jordan and
                  Pieter Abbeel},
  editor       = {Yoshua Bengio and
                  Yann LeCun},
  title        = {High-Dimensional Continuous Control Using Generalized Advantage Estimation},
  booktitle    = {4th International Conference on Learning Representations, {ICLR} 2016,
                  San Juan, Puerto Rico, May 2-4, 2016, Conference Track Proceedings},
  year         = {2016},
  url          = {http://arxiv.org/abs/1506.02438},
  bibsource    = {dblp computer science bibliography, https://dblp.org}
}

@inproceedings{vaswani2017attention,
  author       = {Ashish Vaswani and
                  Noam Shazeer and
                  Niki Parmar and
                  Jakob Uszkoreit and
                  Llion Jones and
                  Aidan N. Gomez and
                  Lukasz Kaiser and
                  Illia Polosukhin},
  editor       = {Isabelle Guyon and
                  Ulrike von Luxburg and
                  Samy Bengio and
                  Hanna M. Wallach and
                  Rob Fergus and
                  S. V. N. Vishwanathan and
                  Roman Garnett},
  title        = {Attention is All you Need},
  booktitle    = {Advances in Neural Information Processing Systems 30: Annual Conference
                  on Neural Information Processing Systems 2017, December 4-9, 2017,
                  Long Beach, CA, {USA}},
  pages        = {5998--6008},
  year         = {2017},
  url          = {https://proceedings.neurips.cc/paper/2017/hash/3f5ee243547dee91fbd053c1c4a845aa-Abstract.html},
  bibsource    = {dblp computer science bibliography, https://dblp.org}
}

@inproceedings{tancik2020fourier,
  author       = {Matthew Tancik and
                  Pratul P. Srinivasan and
                  Ben Mildenhall and
                  Sara Fridovich{-}Keil and
                  Nithin Raghavan and
                  Utkarsh Singhal and
                  Ravi Ramamoorthi and
                  Jonathan T. Barron and
                  Ren Ng},
  editor       = {Hugo Larochelle and
                  Marc'Aurelio Ranzato and
                  Raia Hadsell and
                  Maria{-}Florina Balcan and
                  Hsuan{-}Tien Lin},
  title        = {Fourier Features Let Networks Learn High Frequency Functions in Low
                  Dimensional Domains},
  booktitle    = {Advances in Neural Information Processing Systems 33: Annual Conference
                  on Neural Information Processing Systems 2020, NeurIPS 2020, December
                  6-12, 2020, virtual},
  year         = {2020},
  url          = {https://proceedings.neurips.cc/paper/2020/hash/55053683268957697aa39fba6f231c68-Abstract.html},
  bibsource    = {dblp computer science bibliography, https://dblp.org}
}

@inproceedings{he2016resnet,
  author       = {Kaiming He and
                  Xiangyu Zhang and
                  Shaoqing Ren and
                  Jian Sun},
  title        = {Deep Residual Learning for Image Recognition},
  booktitle    = {2016 {IEEE} Conference on Computer Vision and Pattern Recognition,
                  {CVPR} 2016, Las Vegas, NV, USA, June 27-30, 2016},
  pages        = {770--778},
  publisher    = {{IEEE} Computer Society},
  year         = {2016},
  url          = {https://doi.org/10.1109/CVPR.2016.90},
  doi          = {10.1109/CVPR.2016.90},
  bibsource    = {dblp computer science bibliography, https://dblp.org}
}

@inproceedings{tobin2017domainrand,
  author       = {Josh Tobin and
                  Rachel Fong and
                  Alex Ray and
                  Jonas Schneider and
                  Wojciech Zaremba and
                  Pieter Abbeel},
  title        = {Domain randomization for transferring deep neural networks from simulation
                  to the real world},
  booktitle    = {2017 {IEEE/RSJ} International Conference on Intelligent Robots and
                  Systems, {IROS} 2017, Vancouver, BC, Canada, September 24-28, 2017},
  pages        = {23--30},
  publisher    = {{IEEE}},
  year         = {2017},
  url          = {https://doi.org/10.1109/IROS.2017.8202133},
  doi          = {10.1109/IROS.2017.8202133},
  bibsource    = {dblp computer science bibliography, https://dblp.org}
}

@inproceedings{tan2018simtoreal,
  author       = {Jie Tan and
                  Tingnan Zhang and
                  Erwin Coumans and
                  Atil Iscen and
                  Yunfei Bai and
                  Danijar Hafner and
                  Steven Bohez and
                  Vincent Vanhoucke},
  editor       = {Hadas Kress{-}Gazit and
                  Siddhartha S. Srinivasa and
                  Tom Howard and
                  Nikolay Atanasov},
  title        = {Sim-to-Real: Learning Agile Locomotion For Quadruped Robots},
  booktitle    = {Robotics: Science and Systems XIV, Carnegie Mellon University, Pittsburgh,
                  Pennsylvania, USA, June 26-30, 2018},
  year         = {2018},
  url          = {http://www.roboticsproceedings.org/rss14/p10.html},
  doi          = {10.15607/RSS.2018.XIV.010},
  bibsource    = {dblp computer science bibliography, https://dblp.org}
}

@inproceedings{peng2018dynamicsrand,
  author       = {Xue Bin Peng and
                  Marcin Andrychowicz and
                  Wojciech Zaremba and
                  Pieter Abbeel},
  title        = {Sim-to-Real Transfer of Robotic Control with Dynamics Randomization},
  booktitle    = {2018 {IEEE} International Conference on Robotics and Automation, {ICRA}
                  2018, Brisbane, Australia, May 21-25, 2018},
  pages        = {1--8},
  publisher    = {{IEEE}},
  year         = {2018},
  url          = {https://doi.org/10.1109/ICRA.2018.8460528},
  doi          = {10.1109/ICRA.2018.8460528},
  bibsource    = {dblp computer science bibliography, https://dblp.org}
}

@inproceedings{kumar2021rma,
  author       = {Ashish Kumar and
                  Zipeng Fu and
                  Deepak Pathak and
                  Jitendra Malik},
  editor       = {Dylan A. Shell and
                  Marc Toussaint and
                  M. Ani Hsieh},
  title        = {{RMA:} Rapid Motor Adaptation for Legged Robots},
  booktitle    = {Robotics: Science and Systems XVII, Virtual Event, July 12-16, 2021},
  year         = {2021},
  url          = {https://doi.org/10.15607/RSS.2021.XVII.011},
  doi          = {10.15607/RSS.2021.XVII.011},
  bibsource    = {dblp computer science bibliography, https://dblp.org}
}

@article{ji2022concurrent,
  author       = {Gwanghyeon Ji and
                  Juhyeok Mun and
                  Hyeongjun Kim and
                  Jemin Hwangbo},
  title        = {Concurrent Training of a Control Policy and a State Estimator for
                  Dynamic and Robust Legged Locomotion},
  journal      = {{IEEE} Robotics Autom. Lett.},
  volume       = {7},
  number       = {2},
  pages        = {4630--4637},
  year         = {2022},
  url          = {https://doi.org/10.1109/LRA.2022.3151396},
  doi          = {10.1109/LRA.2022.3151396},
  bibsource    = {dblp computer science bibliography, https://dblp.org}
}

@inproceedings{agarwal2022egocentric,
  author       = {Ananye Agarwal and
                  Ashish Kumar and
                  Jitendra Malik and
                  Deepak Pathak},
  editor       = {Karen Liu and
                  Dana Kulic and
                  Jeffrey Ichnowski},
  title        = {Legged Locomotion in Challenging Terrains using Egocentric Vision},
  booktitle    = {Conference on Robot Learning, CoRL 2022, 14-18 December 2022, Auckland,
                  New Zealand},
  series       = {Proceedings of Machine Learning Research},
  pages        = {403--415},
  publisher    = {{PMLR}},
  year         = {2022},
  url          = {https://proceedings.mlr.press/v205/agarwal23a.html},
  bibsource    = {dblp computer science bibliography, https://dblp.org}
}

@inproceedings{zhuang2023robotparkour,
  author       = {Ziwen Zhuang and
                  Zipeng Fu and
                  Jianren Wang and
                  Christopher G. Atkeson and
                  S{\"{o}}ren Schwertfeger and
                  Chelsea Finn and
                  Hang Zhao},
  editor       = {Jie Tan and
                  Marc Toussaint and
                  Kourosh Darvish},
  title        = {Robot Parkour Learning},
  booktitle    = {Conference on Robot Learning, CoRL 2023, 6-9 November 2023, Atlanta,
                  GA, {USA}},
  series       = {Proceedings of Machine Learning Research},
  pages        = {73--92},
  publisher    = {{PMLR}},
  year         = {2023},
  url          = {https://proceedings.mlr.press/v229/zhuang23a.html},
  bibsource    = {dblp computer science bibliography, https://dblp.org}
}

@inproceedings{loquercio2022crossmodal,
  author       = {Antonio Loquercio and
                  Ashish Kumar and
                  Jitendra Malik},
  title        = {Learning Visual Locomotion with Cross-Modal Supervision},
  booktitle    = {{IEEE} International Conference on Robotics and Automation, {ICRA}
                  2023, London, UK, May 29 - June 2, 2023},
  pages        = {7295--7302},
  publisher    = {{IEEE}},
  year         = {2023},
  url          = {https://doi.org/10.1109/ICRA48891.2023.10160760},
  doi          = {10.1109/ICRA48891.2023.10160760},
  bibsource    = {dblp computer science bibliography, https://dblp.org}
}

@inproceedings{yang2022crossmodaltransformer,
  author       = {Ruihan Yang and
                  Minghao Zhang and
                  Nicklas Hansen and
                  Huazhe Xu and
                  Xiaolong Wang},
  title        = {Learning Vision-Guided Quadrupedal Locomotion End-to-End with Cross-Modal
                  Transformers},
  booktitle    = {The Tenth International Conference on Learning Representations, {ICLR}
                  2022, Virtual Event, April 25-29, 2022},
  publisher    = {OpenReview.net},
  year         = {2022},
  url          = {https://openreview.net/forum?id=nhnJ3oo6AB},
  bibsource    = {dblp computer science bibliography, https://dblp.org}
}

@article{radosavovic2024humanoid,
  author       = {Ilija Radosavovic and
                  Tete Xiao and
                  Bike Zhang and
                  Trevor Darrell and
                  Jitendra Malik and
                  Koushil Sreenath},
  title        = {Real-world humanoid locomotion with reinforcement learning},
  journal      = {Sci. Robotics},
  volume       = {9},
  number       = {89},
  year         = {2024},
  url          = {https://doi.org/10.1126/scirobotics.adi9579},
  doi          = {10.1126/SCIROBOTICS.ADI9579},
  bibsource    = {dblp computer science bibliography, https://dblp.org}
}

@inproceedings{radosavovic2024tokenprediction,
  author       = {Ilija Radosavovic and
                  Bike Zhang and
                  Baifeng Shi and
                  Jathushan Rajasegaran and
                  Sarthak Kamat and
                  Trevor Darrell and
                  Koushil Sreenath and
                  Jitendra Malik},
  editor       = {Amir Globersons and
                  Lester Mackey and
                  Danielle Belgrave and
                  Angela Fan and
                  Ulrich Paquet and
                  Jakub M. Tomczak and
                  Cheng Zhang},
  title        = {Humanoid Locomotion as Next Token Prediction},
  booktitle    = {Advances in Neural Information Processing Systems 37: Annual Conference
                  on Neural Information Processing Systems 2024, NeurIPS 2024, Vancouver,
                  BC, Canada, December 10 - 15, 2024},
  year         = {2024},
  url          = {http://papers.nips.cc/paper\_files/paper/2024/hash/90afd20dc776bc8849c31d61a0763a0b-Abstract-Conference.html},
  bibsource    = {dblp computer science bibliography, https://dblp.org}
}

@article{hao2026cref,
  title         = {{CReF}: Cross-modal and Recurrent Fusion for
                   Depth-conditioned Humanoid Locomotion},
  author        = {Hao, Yuan and Yu, Ruiqi and Luo, Shixin and Zhang, Guoteng
                   and Wu, Jun and Zhu, Qiuguo},
  journal       = {arXiv preprint arXiv:2603.29452},
  year          = {2026},
  eprint        = {2603.29452},
  archivePrefix = {arXiv},
  primaryClass  = {cs.RO},
  url           = {https://arxiv.org/abs/2603.29452}
}

@article{luo2026lf2wb,
  title         = {Look Forward to Walk Backward: Efficient Terrain Memory
                   for Backward Locomotion with Forward Vision},
  author        = {Luo, Shixin and Li, Songbo and Hao, Yuan and Wang, Yaqi
                   and Zheng, Jun and Wu, Jun and Zhu, Qiuguo},
  journal       = {arXiv preprint arXiv:2603.03138},
  year          = {2026},
  eprint        = {2603.03138},
  archivePrefix = {arXiv},
  primaryClass  = {cs.RO},
  url           = {https://arxiv.org/abs/2603.03138}
}

%===============================================================================
\appendix

% --- Appendix-specific figure and table numbering ---
\setcounter{figure}{1}
\setcounter{table}{1}
\renewcommand{\thefigure}{A\arabic{figure}}
\renewcommand{\thetable}{A\arabic{table}}
\providecommand{\theHfigure}{}
\providecommand{\theHtable}{}
\renewcommand{\theHfigure}{appendix.figure.\arabic{figure}}
\renewcommand{\theHtable}{appendix.table.\arabic{table}}

\section*{Appendix}

App.~\ref{app:eval-protocols} specifies the shared stress-test environment
and reconstruction metrics, including the matched offline reconstructor
ablation in App.~\ref{app:offline-recon}.  App.~\ref{app:arch}
lists dimensions for the action head, student LSTM bypass and QR.
App.~\ref{app:training} gives all training hyperparameters, including PPO
and reconstructor optimization, AMP, task reward, terrain curriculum and
domain randomization.  App.~\ref{app:platform} describes the real-robot
platform, runtime configuration, and indoor course layout.

%===============================================================================
\section{Evaluation Protocols}
\label{app:eval-protocols}

\subsection{Stress-Test Environment and Commands}
\label{app:stress-env}

The reconstruction and locomotion-quality evaluations in
\S\ref{sec:exp-recon}--\S\ref{sec:exp-e2e} share a common stress-test
protocol.  For each reported terrain, we run a separate evaluation job
containing one terrain type at one fixed curriculum difficulty.  This
avoids cross-terrain interference from mixing different terrain
families or difficulty levels in the same scene.  For the
highest-difficulty results, the difficulty is fixed to $0.99$, and we
run $N=100$ rollouts.  Observation noise and external pushes are
disabled; heading commands are fixed to zero, lateral velocity
commands are zero, and the forward command is sampled from
$v_x\in[0.4,1.0]$\,m/s.  Episodes have a $30$\,s time limit for all methods.

\paragraph{Rollout termination.}
Each robot starts at its assigned terrain origin and runs until success or
failure.  Success requires root translation beyond
$d_\mathrm{pass}=5.0$\,m along positive $x$ relative to the origin; an
environment reset or the time limit also ends the rollout.  Ended rollouts
are removed from subsequent statistics, preventing dilution by
post-termination padding in the reported metrics.

\subsection{Terrain-Reconstructor Evaluation}
\label{app:terrain-recon-eval}

We evaluate terrain reconstruction inside the same long-horizon rollout
used for the locomotion stress test (App.~\ref{app:stress-env}).  Each
reconstructor is evaluated through its corresponding distilled student
policy.  The teacher, controller architecture, terrain configuration,
command distribution, training budget, and distillation objective are held
fixed across methods, and reconstruction error is measured on the states
visited by each student under its closed-loop visitation distribution.

\paragraph{Reconstruction metric.}
At each active control step, we record the simulator ground-truth map
$m_{i,t}$ and read estimate $\hat m_{i,t}$ from the policy acting on
deployable observations.  For rollout $i$ with $T_i$ active steps, we
compute its per-step height-map L1 error as
\begin{equation}
e_i^m=\frac{1}{T_i}\sum_{t=1}^{T_i}
\frac{1}{|\mathcal G|}\|\hat m_{i,t}-m_{i,t}\|_1,
\label{eq:terrain-recon-eval-metrics}
\end{equation}
Here, $\mathcal G$ is the $16{\times}32$ height-map grid, and each reported
terrain error averages $e_i^m$ over $N$ rollouts.

\subsection{Matched Offline Reconstructor Comparison}
\label{app:offline-recon}

The main experiments train each reconstructor online with its corresponding
student policy.  This coupling supplies a larger and more diverse stream of
curriculum states, aligns reconstruction supervision with the visitation
distribution of the deployed policy, and directly measures how reconstruction
quality affects closed-loop policy deployment.  Because the sampling policy
also determines which states each reconstructor observes, we add a matched
offline comparison to isolate reconstructor architecture from the sampling
policy using one fixed sampler, a common dataset, and identical optimization.
A fixed teacher sampler collects approximately 249k labeled frames over four
terrain seeds.  QR, START~\citep{yu2026start}, and DPL~\citep{sun2025dpl}
receive identical data, output-only L1 supervision, and 15k optimization
updates.  Evaluation covers 850 unseen terrain instances in five independently
generated held-out sets; Table~\ref{tab:offline-recon} reports mean$\pm$std
across the five sets.  Edge F1@1-cell is computed on the 582 instances
containing ground-truth edges across the matched held-out comparison.

\begin{table*}[!htbp]
\centering
\caption{\textbf{Matched offline reconstruction comparison.}  Aggregate and
level-9 height-map L1 are reported in cm ($\downarrow$); Edge F1@1-cell is
higher-is-better under identical supervision and optimization.}
\label{tab:offline-recon}
{\scriptsize
\setlength{\tabcolsep}{3.2pt}
\renewcommand{\arraystretch}{1.05}
\newcommand{\mstd}[2]{\shortstack{#1\\[-0.25em]{\tiny$\pm #2$}}}
\newcommand{\bmstd}[2]{\shortstack{\textbf{#1}\\[-0.25em]{\tiny$\pm #2$}}}
\begin{tabular*}{\textwidth}{@{\extracolsep{\fill}}lcccccccc@{}}
\toprule
\multirow{2}{*}{Method} & \multirow{2}{*}{\#P} &
\multicolumn{2}{c}{Aggregate (cm)$\downarrow$} & Edge$\uparrow$ &
\multicolumn{4}{c}{Level-9 terrain (cm)$\downarrow$} \\
\cmidrule(lr){3-4}\cmidrule(lr){6-9}
& & Overall & Hard trav. & F1@1 & Down-25 & Up-25 & Stones & Rough \\
\midrule
QR (ours) & \textbf{0.54M} & \bmstd{2.98}{0.07} & \bmstd{3.93}{0.07}
& \bmstd{.386}{.008} & \bmstd{5.25}{0.89} & \bmstd{4.91}{0.68}
& \bmstd{10.41}{1.80} & \bmstd{3.70}{0.21} \\
START~\citep{yu2026start} & 0.62M & \mstd{4.32}{0.13} & \mstd{5.62}{0.15}
& \mstd{.242}{.004} & \mstd{8.91}{2.18} & \mstd{6.78}{0.59}
& \mstd{12.87}{1.97} & \mstd{4.77}{0.29} \\
DPL~\citep{sun2025dpl} & 0.78M & \mstd{4.36}{0.10} & \mstd{5.57}{0.11}
& \mstd{.214}{.004} & \mstd{7.88}{1.92} & \mstd{6.64}{0.76}
& \mstd{13.11}{2.55} & \mstd{4.71}{0.22} \\
\bottomrule
\end{tabular*}
}
\end{table*}

QR lowers overall height-map L1 by 30.9\% versus START and 31.6\% versus
DPL.  On the edge-containing instances, F1@1-cell increases by 59.4\% and
79.9\%, respectively.  QR also records the lowest L1 on the hard-traversability
aggregate and all four reported level-9 terrain groups.

\begin{figure*}[!htbp]
\centering
\includegraphics[width=0.96\textwidth]{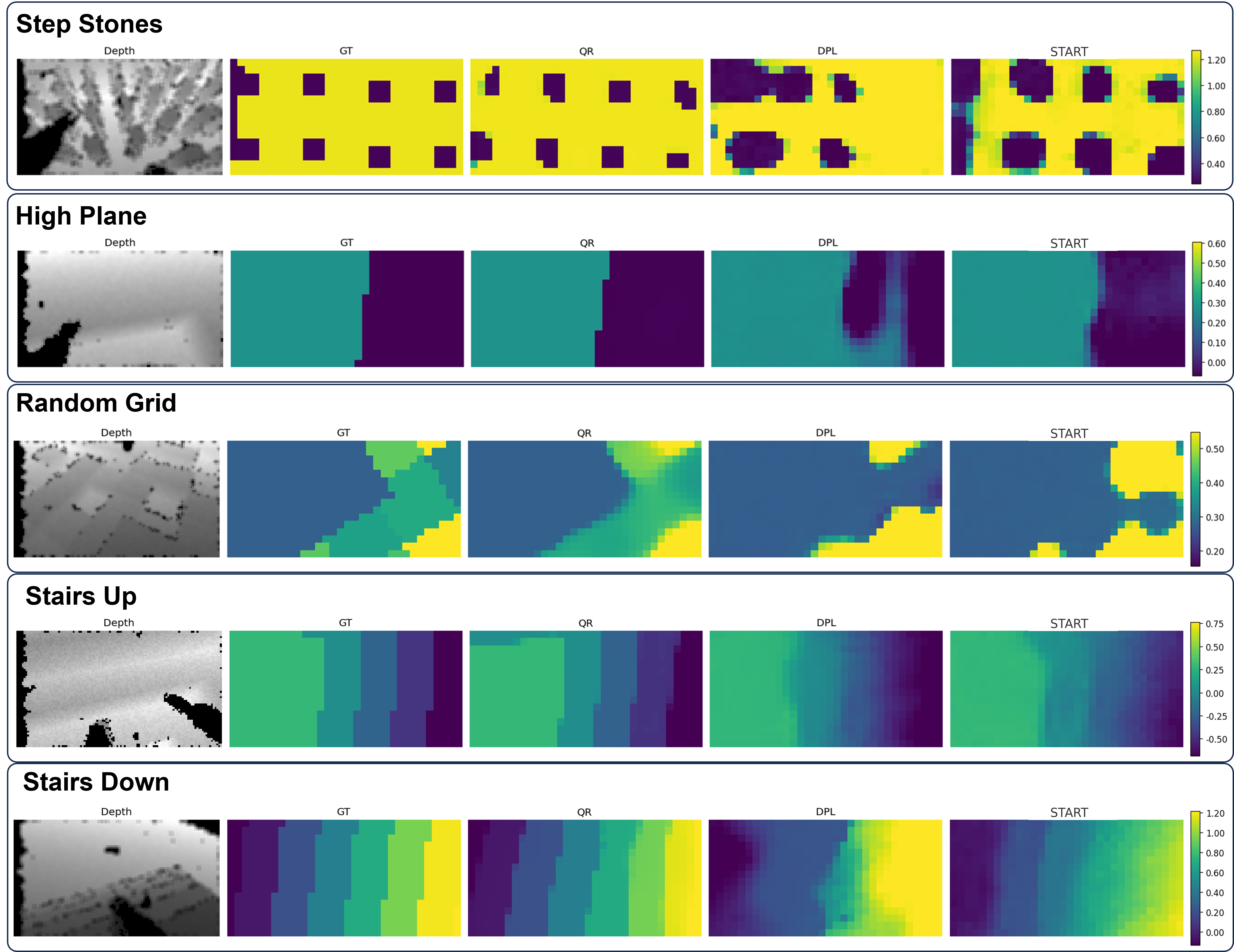}
\caption{\textbf{Qualitative height-map reconstruction at difficulty
$0.99$.}  Rows show Step Stones, High Plane, Random Grid and Stairs Up/Down;
columns show the latest depth, privileged ground truth, QR, DPL and START.
Each row uses its own color scale.  QR preserves sharper discontinuities,
especially on Step Stones, where footholds remain separated and
aligned with ground truth.}
\label{fig:height-map-qualitative}
\end{figure*}

\subsection{Seed-Paired Reduced-Compute TA-MSE Ablation}
\label{app:reduced-compute-tamse}

We evaluate training-seed variation under a matched reduced-compute protocol
using half-resolution depth and $2\!\times\!2048$ environments.  Starting
from seed-matched D-PPO checkpoints (seeds 101, 202, and 303), we attach QR
and continue training for 3,000 iterations with either MSE+PPO or \TAMSE.
Evaluation covers ten terrain levels with 64 episodes for every
method--seed--terrain--level cell.  Easy, medium, and hard aggregate levels
denote L0--3, L4--6, and L7--9, respectively.

\begin{table*}[!htbp]
\centering
\caption{\textbf{Three-seed results under the reduced-compute continuation
protocol.} Values are mean$\pm$std across the three seed-paired
runs; $\Delta$ is computed per seed before aggregation.}
\label{tab:reduced-compute-tamse}
\scriptsize
\setlength{\tabcolsep}{3.2pt}
\begin{tabular}{@{}lccccccc@{}}
\toprule
& \multicolumn{3}{c}{Aggregate success (\%)} &
\multicolumn{4}{c}{Level-9 success (\%)} \\
\cmidrule(lr){2-4}\cmidrule(lr){5-8}
Method & Easy & Medium & Hard & Up-25 & Down-25 & Rough & Stones \\
\midrule
MSE+PPO & $99.23\!\pm\!0.27$ & $98.23\!\pm\!0.26$ & $97.14\!\pm\!0.67$ &
$97.92\!\pm\!3.61$ & $100.00\!\pm\!0.00$ & $96.35\!\pm\!1.80$ & $72.66\!\pm\!5.12$ \\
\TAMSE & $\mathbf{99.24\!\pm\!0.15}$ & $\mathbf{98.39\!\pm\!0.15}$ & $\mathbf{97.45\!\pm\!0.31}$ &
$\mathbf{98.96\!\pm\!0.90}$ & $97.92\!\pm\!3.61$ & $\mathbf{96.88\!\pm\!3.13}$ & $\mathbf{79.43\!\pm\!5.97}$ \\
$\Delta$ (pp) & $+0.02\!\pm\!0.40$ & $+0.15\!\pm\!0.40$ & $+0.31\!\pm\!0.66$ &
$+1.04\!\pm\!3.25$ & $-2.08\!\pm\!3.61$ & $+0.52\!\pm\!3.61$ & $\mathbf{+6.77\!\pm\!5.86}$ \\
\bottomrule
\end{tabular}
\end{table*}

\TAMSE{} improves level-9 stepping-stone success from
$72.66\%$ to $79.43\%$ ($+6.77$ percentage points, seed-paired), while easy
and medium terrain remain near saturation for both objectives.

\FloatBarrier

%===============================================================================
\section{Network Layer Dimensions}
\label{app:arch}

\begin{figure}[!htbp]
\centering
\includegraphics[width=0.6\linewidth]{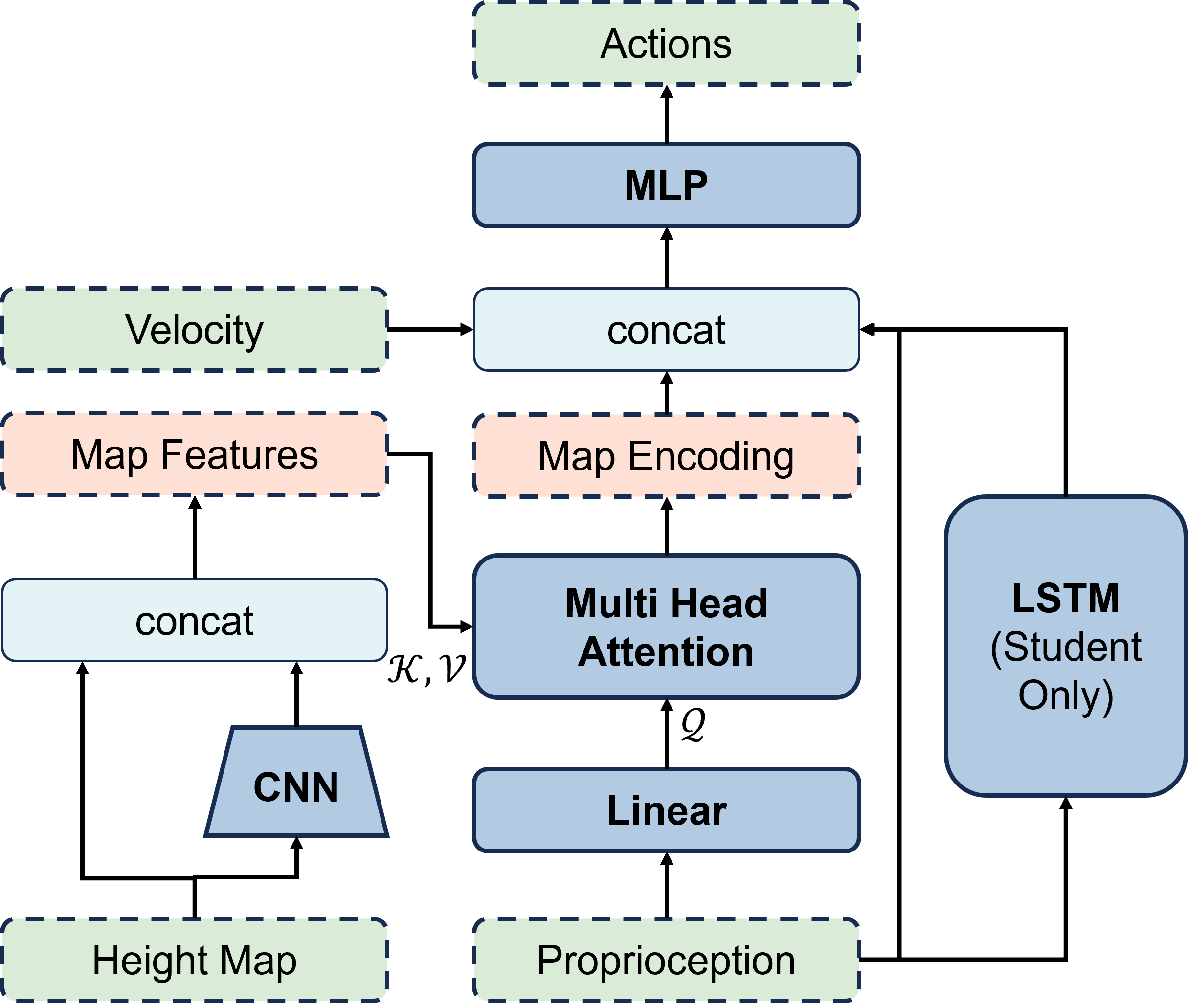}
\caption{\textbf{Expanded \AME{} teacher--student policy architecture from
Figure~\ref{fig:framework}.}  The teacher and student share the height-map and
proprioceptive encoding pathway; the student additionally uses the LSTM
proprioceptive bypass before producing each deployed action on the robot.}
\label{fig:ame-policy-detail}
\end{figure}

Figure~\ref{fig:ame-policy-detail} expands the \AME{} policy block in
Figure~\ref{fig:framework}, showing the shared height-map and proprioceptive
encoding pathway and the student-only LSTM bypass.  Table~\ref{tab:arch-dims}
lists the corresponding action-head, LSTM-bypass, and QR dimensions.  Map,
depth, and temporal dimensions follow the deployment configuration in
App.~\ref{app:platform-hw} during onboard inference in all hardware trials.

\begin{table}[!htbp]
\centering
\caption{\textbf{Layer-wise feature dimensions} for the action head,
the student LSTM bypass and the Query Reconstructor (QR).  Dimensions
reflect the deployable observation used at run time.}
\label{tab:arch-dims}
\small
\setlength{\tabcolsep}{4pt}
\begin{tabular}{@{}lll@{}}
\toprule
Module & Component & Setting \\
\midrule
\multicolumn{3}{@{}l}{\emph{Action head (MLP)}} \\
Teacher action head & Hidden dims & $[256,256,128]$ \\
Student action head & Hidden dims & $[512,256,128]$ \\
Activation & --- & ELU \\
\midrule
\multicolumn{3}{@{}l}{\emph{Student LSTM bypass}} \\
Input & Single-frame proprio & dim $84$ \\
Recurrence & LSTM hidden / layers & $128$ / $1$ \\
Output fusion & Concatenated with encoder & $128$ \\
\midrule
\multicolumn{3}{@{}l}{\emph{Query Reconstructor (QR)}} \\
Map queries & Number, grid & $512$ ($16{\times}32$) \\
Position encoding & Fourier frequencies & $12$ \\
Memory transformer & $d_\text{model}$, heads & $64$, $8$ \\
Self-attention layers & Memory encoder & $4$ \\
Cross-attention layers & Query decoder & $2$ \\
Depth window & History length, stride & $4$ frames, $4$ \\
Image input & Depth resolution & $113{\times}64$ \\
Proprio history & Encoder window & $10$ frames \\
\bottomrule
\end{tabular}
\end{table}

%===============================================================================
\section{Training Details}
\label{app:training}

\subsection{Optimization Hyperparameters}
\label{app:hp}

Table~\ref{tab:optimization-hp} lists the PPO~\citep{schulman2017ppo} and
reconstructor optimizer settings used across the three-stage training
schedule described in \S\ref{sec:overview}; the GAE
parameter $\lambda$ directly follows~\citet{schulman2016gae}.

\begin{table}[!htbp]
\centering
\caption{\textbf{Optimization hyperparameters} for PPO and QR in the
three-stage curriculum.}
\label{tab:optimization-hp}
\small
\setlength{\tabcolsep}{5pt}
\begin{tabular}{@{}lll@{}}
\toprule
Component & Hyperparameter & Value \\
\midrule
PPO & Learning rate & $1.0{\times}10^{-3}$ \\
PPO & Learning-rate schedule & Adaptive \\
PPO & Target KL divergence & $0.01$ \\
PPO & Clipping ratio $\epsilon$ & $0.2$ \\
PPO & Discount factor $\gamma$ & $0.99$ \\
PPO & GAE parameter $\lambda$ & $0.95$ \\
PPO & Value loss coefficient & $1.0$ \\
PPO (teacher) & Entropy coefficient & $0.005$ \\
PPO (student) & Entropy coefficient & $0.002$ \\
PPO & Maximum gradient norm & $1.0$ \\
PPO & Learning epochs per update & $5$ \\
PPO & Mini-batches per update & $4$ \\
PPO & Rollout length & $24$ steps/env \\
\midrule
Distillation & Auxiliary MSE weight $\lambda$ & $5$ \\
Distillation & Reward-channel MSE weight $\beta$ & $1$ \\
\midrule
Reconstructor & Learning rate & $5.0{\times}10^{-4}$ \\
Reconstructor & Learning-rate schedule & Cosine \\
\bottomrule
\end{tabular}
\end{table}

\subsection{Training Budget and Onboard Runtime}
\label{app:compute}

Training uses four H800 GPUs.  Table~\ref{tab:training-budget} gives
environments/GPU, updates, and wall-clock time.

\begin{table}[!htbp]
\centering
\caption{\textbf{Training budget} for the three-stage SOLO pipeline.}
\label{tab:training-budget}
\small
\setlength{\tabcolsep}{4pt}
\begin{tabular}{@{}lrrr@{}}
\toprule
Stage & Envs/GPU & Iterations & Time (h) \\
\midrule
Teacher PPO & 4096 & 50,000 & $\sim$90 \\
Privileged distillation & 4096 & 20,000 & $\sim$64 \\
QR fine-tuning & 2048 & 5,000 & $\sim$20 \\
\bottomrule
\end{tabular}
\end{table}

On a Jetson AGX Thor in MAXN mode, 1,000 batch-1 TensorRT inferences with
live 60-Hz D455 depth take $1.03\pm0.21$\,ms (p95: $1.59$\,ms).  The full
observation-build, inference, and action-postprocessing pipeline takes
$1.15\pm0.23$\,ms on average, within the 20-ms budget of the 50-Hz controller.
The deployed TensorRT engine occupies 7.56\,MB during onboard operation.

\subsection{Adversarial Motion Prior}
\label{app:amp}

We use AMP~\citep{peng2021amp} as a motion-style regularizer during both
teacher pretraining and student distillation.  The reference motion set
contains medium- and low-speed walking clips and in-place turning
clips.  Table~\ref{tab:amp-hp} summarizes the AMP parameters used in
all of our training runs.

\begin{table}[!htbp]
\centering
\caption{\textbf{AMP parameters.}  $c_\mathrm{amp}$
weights AMP reward; $\eta$ blends task/AMP rewards.}
\label{tab:amp-hp}
\small
\setlength{\tabcolsep}{4pt}
\begin{tabular}{@{}ll@{}}
\toprule
Parameter & Value \\
\midrule
AMP reward coefficient $c_{\mathrm{amp}}$ & $0.3$ \\
Task-reward interpolation $\eta$ & $0.7$ \\
Discriminator hidden dimensions & $[1024,512]$ \\
\bottomrule
\end{tabular}
\end{table}

\subsection{Task Reward Formulation}
\label{app:reward}

The locomotion reward is $r_t=\sum_k w_k r_k$; Table~\ref{tab:task-reward}
lists its sparse-foothold terms and weights.  Here, $\mathbf v^{c}$ and
$\omega^{c}$ are commanded velocities; superscripts $b$, $w$, and $y$
denote body, world, and yaw-aligned frames; and $\mathbf g_{xy}^{B}$ is
gravity's horizontal component in body frame $B$.  The foot set
$\mathcal F=\{L,R\}$ includes only the left and right ankle-roll contact
links during locomotion training.

\begin{table*}[t]
\centering
\caption{\textbf{Task reward terms used for sparse-foothold
locomotion.}  Indicator functions are denoted by
$\mathbf{1}[\cdot]$; $c_f$ indicates contact of foot $f$, and
$\operatorname{clip}(x,l,u)$ clamps $x$ within $[l,u]$.}
\label{tab:task-reward}
\scriptsize
\renewcommand{\arraystretch}{1.10}
\setlength{\tabcolsep}{3pt}
\begin{tabular}{@{}p{0.27\textwidth}p{0.59\textwidth}r@{}}
\toprule
Term & Definition $r_k$ & $w_k$ \\
\midrule
\multicolumn{3}{@{}l}{\emph{Command tracking and regularization}} \\
\texttt{track\_lin\_vel\_xy\_exp}
  & $\exp\!\left(-\|\mathbf v_{xy}^{c}-\mathbf v_{xy}^{y}\|_2^2/0.5^2\right)$ & $3.0$ \\
\texttt{track\_ang\_vel\_z\_exp}
  & $\exp\!\left(-(\omega_z^{c}-\omega_z^{w})^2/0.5^2\right)$ & $5.0$ \\
\texttt{lin\_vel\_z\_l2}
  & $(v_z^{b})^2$ & $-0.25$ \\
\texttt{ang\_vel\_xy\_l2}
  & $\|\boldsymbol{\omega}_{xy}^{b}\|_2^2$ & $-0.05$ \\
\texttt{energy}
  & $\|\boldsymbol{\tau}\odot\dot{\mathbf q}\|_2$ & $-1.0{\times}10^{-3}$ \\
\texttt{dof\_acc\_l2}
  & $\|\ddot{\mathbf q}\|_2^2$ & $-2.5{\times}10^{-7}$ \\
\texttt{action\_rate\_l2}
  & $\|\mathbf a_t-\mathbf a_{t-1}\|_2^2$ & $-0.01$ \\
\texttt{dof\_pos\_limits}
  & $\sum_j \max(q_j-q_j^{\max},\,q_j^{\min}-q_j,\,0)$ & $-2.0$ \\
\texttt{joint\_deviation\_hip}
  & $\|\mathbf q_{\mathrm{hip\ yaw,roll}}-\mathbf q_{0,\mathrm{hip\ yaw,roll}}\|_1$ & $-0.15$ \\
\texttt{joint\_deviation\_arms}
  & $\|\mathbf q_{\mathrm{waist,shoulder,elbow}}-\mathbf q_{0,\mathrm{waist,shoulder,elbow}}\|_1$ & $-0.15$ \\
\texttt{joint\_deviation\_legs}
  & $\|\mathbf q_{\mathrm{hip\ pitch,knee,ankle}}-\mathbf q_{0,\mathrm{hip\ pitch,knee,ankle}}\|_1$ & $-0.02$ \\
\midrule
\multicolumn{3}{@{}l}{\emph{Stability and contact}} \\
\texttt{undesired\_contacts}
  & $\sum_{b\notin\mathcal F}\mathbf{1}[\|\mathbf F_b\|_2>1]$ & $-1.0$ \\
\texttt{fly}
  & $\mathbf{1}[\sum_{f\in\mathcal F}c_f=0]$ & $-3.0$ \\
\texttt{body\_orientation\_l2}
  & $\|\mathbf g_{xy}^{\mathrm{waist}}\|_2^2$ & $-5.0$ \\
\texttt{flat\_orientation\_l2}
  & $\|\mathbf g_{xy}^{\mathrm{base}}\|_2^2$ & $-1.0$ \\
\texttt{termination\_penalty}
  & $\mathbf{1}[\text{any waist or head link contacts the environment}]$ & $-200.0$ \\
\texttt{feet\_air\_time}
  & $\mathbf{1}_{\mathrm{move}}\mathbf{1}_{\mathrm{single}}\,
     \min(\min(t_L,t_R),0.4)$ & $3.0$ \\
\texttt{feet\_slide}
  & $\sum_{f\in\mathcal F}c_f\|\mathbf v_{xy}^{f}\|_2$ & $-0.25$ \\
\texttt{feet\_force}
  & $\operatorname{clip}(\|\mathbf F_{z}^{\mathcal F}\|_2-500,0,400)$ & $-1.0{\times}10^{-2}$ \\
\texttt{feet\_stumble}
  & $\mathbf{1}[\exists f\in\mathcal F:\|\mathbf F_{xy}^{f}\|_2>|F_z^f|]$ & $-5.0$ \\
\texttt{foothold\_grad\_penalty}
  & $\mathbf{1}[v_x^b>0]\sum_{f\in\mathcal F}c_f
     \left(0.7(d_f^{\mathrm{front}}+d_f^{\mathrm{rear}})
     +0.3(d_f^{\mathrm{left}}+d_f^{\mathrm{right}})\right)$ & $-5.0$ \\
\bottomrule
\end{tabular}
\end{table*}

\paragraph{Foothold gradient penalty.}
For each contacting foot, we finite-difference the local height samples
beneath the ankle-roll link.  Each $d_f^{s}$ is one if any signed height
difference on side $s$ crosses $0.05$\,m and zero otherwise.  The term
penalizes a foot placed across a height discontinuity while the robot is
progressing forward, with higher weight on front/rear edges than on
lateral edges.

\subsection{Terrain Curriculum}
\label{app:terrain}

Training uses procedurally generated terrain patches arranged in a
$10{\times}17$ curriculum grid.  Each patch covers $8.0{\times}8.0$\,m,
with horizontal and vertical resolutions of $0.05$\,m and $0.005$\,m,
respectively, a $20$\,m outer border, and a slope threshold of $0.75$.
Terrain difficulty increases across the ten rows.  The 17 active variants
have equal sampling weight (one column each).  Table~\ref{tab:terrain-curriculum}
groups them by terrain family and reports each range from its easiest to
most difficult level in our experiments.

\begin{table*}[!htbp]
\centering
\caption{\textbf{Procedural terrain curriculum.}  Each family spans its
easiest-to-hardest variants; Share is its normalized fraction of terrain
columns after grouping configurations from that family.}
\label{tab:terrain-curriculum}
\scriptsize
\renewcommand{\arraystretch}{1.12}
\setlength{\tabcolsep}{3pt}
\begin{tabular}{@{}>{\raggedright\arraybackslash}p{0.22\textwidth}c%
>{\raggedright\arraybackslash}p{0.57\textwidth}r@{}}
\toprule
Terrain family & Variants & Curriculum parameters & Share \\
\midrule
Flat ground
  & $1$ & Planar terrain patch & $5.9\%$ \\
Ascending/descending stairs
  & $3+3$ & Step height $0.00{\rightarrow}0.25$\,m; tread width
  $\{0.25,0.30,0.35\}$\,m; platform width $2.5$\,m & $35.3\%$ \\
Ascending/descending slopes
  & $1+1$ & Slope magnitude $0.00{\rightarrow}0.50$; platform width
  $2.5$\,m & $11.8\%$ \\
Boxes and pits
  & $1+1$ & Box height / pit depth $0.00{\rightarrow}0.40$\,m; two
  successive obstacles; platform width $2.5$\,m & $11.8\%$ \\
Random grid
  & $1$ & Cell width $0.45$\,m; height
  $0.00{\rightarrow}0.20$\,m; platform width $2.5$\,m & $5.9\%$ \\
Random rough ground
  & $2$ & Height range $[-0.10,0.10]$\,m; quantization step
  $\{0.05,0.02\}$\,m; downsampling scale $\{0.50,0.20\}$\,m
  & $11.8\%$ \\
Gap
  & $1$ & Gap width $0.10{\rightarrow}0.50$\,m; depth
  $0.10{\rightarrow}1.00$\,m; platform width $2.5$\,m & $5.9\%$ \\
Stepping stones
  & $2$ & Stone width $0.40{\rightarrow}0.20$\,m; inter-stone gap
  $0.20{\rightarrow}0.30$\,m; hole depth
  $0.40{\rightarrow}1.00$\,m; platform width $2.5$\,m & $11.8\%$ \\
\bottomrule
\end{tabular}
\end{table*}

\subsection{Domain Randomization}
\label{app:dr}

\begin{figure*}[!htbp]
\centering
\includegraphics[width=0.96\textwidth]{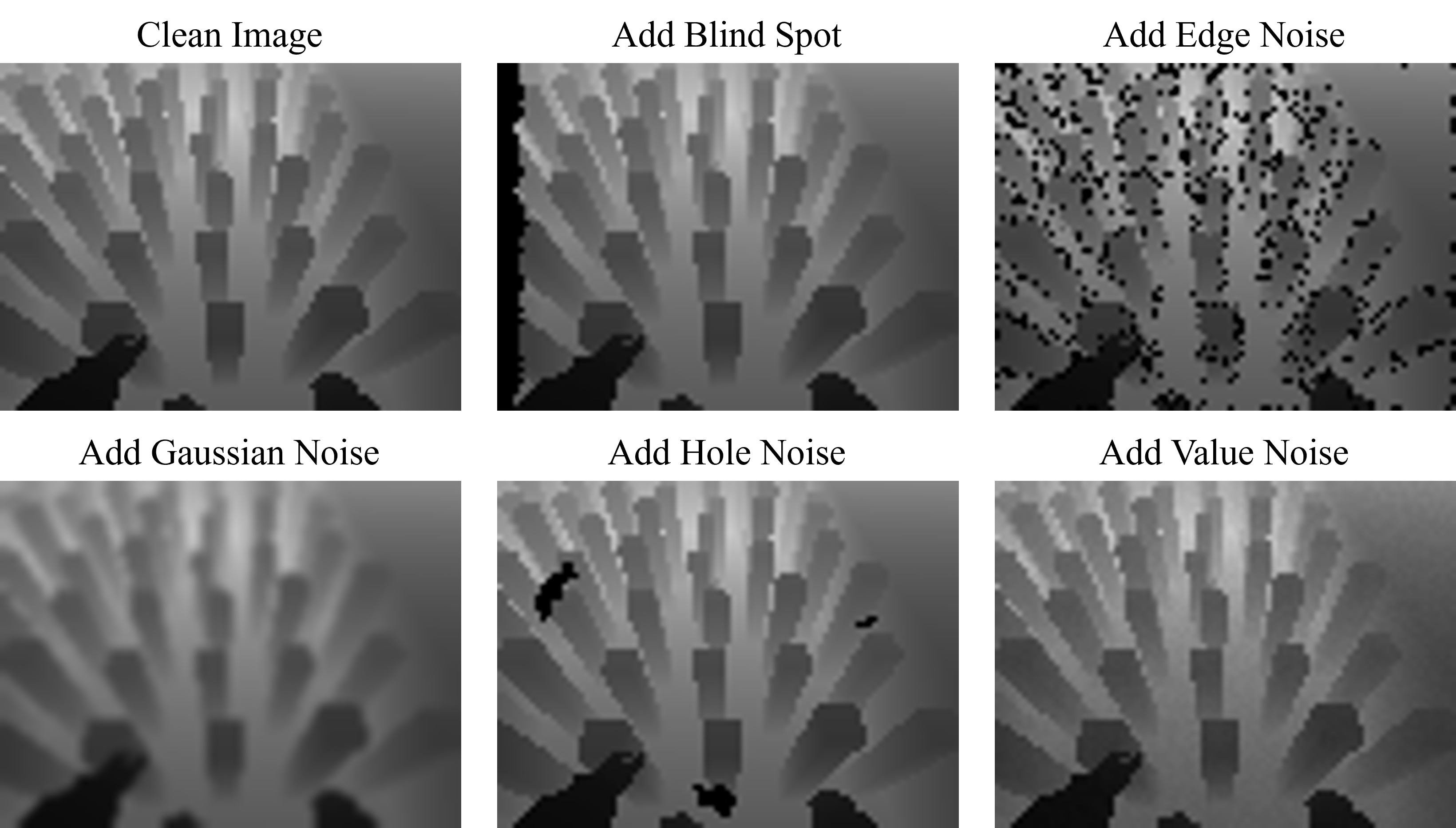}
\caption{\textbf{Synthetic depth-image corruptions} applied during
perceptual policy training.  Samples show randomized value/edge noise,
missing pixels, blind regions, and blur that jointly approximate the
artifact distribution observed from the chest-mounted RealSense D455 at
deployment.}
\label{fig:depth-noise}
\end{figure*}

Table~\ref{tab:domain-rand} summarizes the physical and sensory
randomization~\citep{tobin2017domainrand, peng2018dynamicsrand,
tan2018simtoreal} applied during training, and Figure~\ref{fig:depth-noise}
visualizes the synthetic depth-image corruptions applied to the
perceptual input.  We use $\mathcal U(a,b)$ for uniform sampling and
$\operatorname{LogU}(a,b)$ for log-uniform multiplicative scaling.

\begin{table*}[!htbp]
\centering
\caption{\textbf{Domain randomization parameters} used during training.
$\Delta$ denotes an additive perturbation; sensor-artifact
probabilities are resampled per episode during training.}
\label{tab:domain-rand}
\scriptsize
\renewcommand{\arraystretch}{1.12}
\setlength{\tabcolsep}{3pt}
\begin{tabular}{@{}>{\raggedright\arraybackslash}p{0.19\textwidth}%
>{\raggedright\arraybackslash}p{0.30\textwidth}%
>{\raggedright\arraybackslash}p{0.45\textwidth}@{}}
\toprule
Category & Quantity & Randomization range / setting \\
\midrule
\multicolumn{3}{@{}l}{\emph{Dynamics and initialization}} \\
Contact material
  & Static friction; dynamic friction; restitution
  & $\mathcal U(0.6,1.0)$; $\mathcal U(0.4,0.8)$;
    $\mathcal U(0.0,0.005)$ \\
Rigid-body inertial properties
  & Waist mass offset $\Delta m$
  & $\mathcal U(-3,3)$\,kg \\
Rigid-body inertial properties
  & Waist CoM offset $\Delta x,\Delta y,\Delta z$
  & $\mathcal U(-0.05,0.01)$, $\mathcal U(-0.03,0.03)$,
    $\mathcal U(-0.05,0.05)$\,m \\
Actuator gains
  & Stiffness and damping multipliers
  & $\operatorname{LogU}(0.7,1.3)$ \\
Base initialization
  & Position $(x,y)$; orientation $(\phi,\theta,\psi)$
  & $\mathcal U(-0.5,0.5)$\,m;
    $\mathcal U((-0.25,-0.25,-\pi),(0.25,0.25,\pi))$\,rad \\
Base initialization
  & Linear and angular velocity
  & $\mathcal U(-0.5,0.5)$\,m/s per axis;
    $\mathcal U(-0.5,0.5)$\,rad/s per axis \\
Joint initialization
  & Position scaling about nominal pose; velocity
  & $\mathcal U(0.5,1.5)\,\mathbf q_0$; $0$\,rad/s \\
External push
  & Planar base velocity perturbation
  & Every $\mathcal U(10,15)$\,s;
    $(v_x,v_y)\sim\mathcal U((-1,-1),(1,1))$\,m/s \\
\midrule
\multicolumn{3}{@{}l}{\emph{Observation and perception}} \\
Proprioception
  & Angular velocity; projected gravity; joint position; joint velocity
  & Additive uniform noise scales:
    $\pm0.20$, $\pm0.05$, $\pm0.01$, $\pm1.50$, respectively \\
Camera extrinsics
  & Position offset $\Delta x,\Delta y,\Delta z$
  & $\mathcal U(-0.005,0.005)$, $\mathcal U(-0.010,0.010)$,
    $\mathcal U(-0.010,0.010)$\,m \\
Camera extrinsics
  & Rotation offset $\Delta\phi,\Delta\theta,\Delta\psi$
  & $\mathcal U(-0.05,0.05)$\,rad per axis \\
Depth latency
  & Buffered depth-frame delay & $\mathcal U\{0,1,2,3\}$ frames \\
Depth image corruption
  & Synthetic sensor artifacts
  & Randomized value/edge noise, missing pixels, blind regions, and blur;
    see Fig.~\ref{fig:depth-noise} \\
\bottomrule
\end{tabular}
\end{table*}

\section{Real-Robot Deployment Setup}
\label{app:platform}

\subsection{Hardware Platform and Sensor Configuration}
\label{app:platform-hw}

Our hardware platform is the Omni humanoid.  The policy directly
controls 25 joints: two six-DoF legs (hip yaw/roll/pitch, knee pitch,
and ankle pitch/roll), a three-DoF waist (yaw/roll/pitch), and two
five-DoF arms (shoulder yaw/roll/pitch and elbow yaw/pitch).  A
chest-mounted Intel RealSense D455 provides egocentric depth input
for the perceptive student.  Table~\ref{tab:deployment-platform}
centralizes the hardware, runtime-input and policy-execution configuration
used for all real-robot deployments.

\begin{table}[!htbp]
\centering
\caption{\textbf{Omni deployment and runtime-input configuration.}
The network depth input is downsampled from the raw RealSense stream
and buffered into a short multi-frame history with stochastic delay.  No
external localization or mapping system is used during deployment.}
\label{tab:deployment-platform}
\small
\setlength{\tabcolsep}{4pt}
\begin{tabular}{@{}>{\raggedright\arraybackslash}p{0.34\linewidth}%
>{\raggedright\arraybackslash}p{0.58\linewidth}@{}}
\toprule
Item & Setting \\
\midrule
Platform & Omni humanoid; 25 policy-controlled joints \\
Policy execution & TensorRT engine at 50\,Hz \\
External state input & None; no motion capture, external odometry, or external mapping \\
Depth sensor & Chest-mounted Intel RealSense D455 \\
Raw depth stream & $640{\times}480$ at 60\,Hz \\
Network depth input & $113{\times}64$ resized image \\
Temporal depth input & 4 frames; stride 4 \\
Proprioceptive input & 84-D per frame; 10-frame encoder window \\
Reconstructed state & $16{\times}32$ height map at $0.05$\,m/cell; 3-DoF base velocity \\
\bottomrule
\end{tabular}
\end{table}

\subsection{Indoor Mixed-Terrain Course}
\label{app:platform-indoor}

One uninterrupted run traverses 15\,cm-high stairs, 20\,cm and 30\,cm square
stepping stones, a 45\,cm-wide gap, and a movable obstacle in the continuous
hardware deployment shown in Figure~\ref{fig:indoor-mixed-deployment}.

\begin{figure*}[!htbp]
\centering
\includegraphics[width=0.98\textwidth]{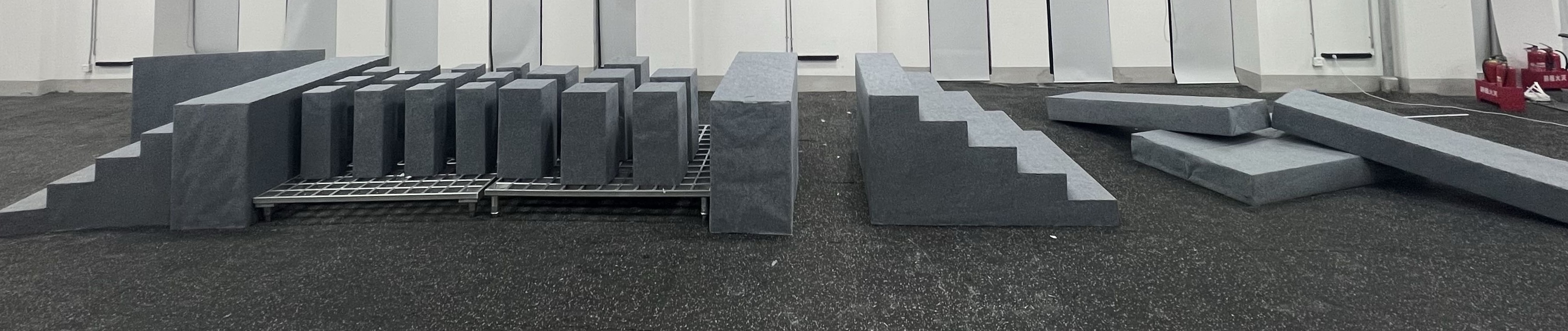}
\caption{\textbf{Indoor mixed-terrain course layout.}  The continuous course
combines 15\,cm-high stairs, 20\,cm and 30\,cm square stepping stones, a
45\,cm-wide gap, and a movable obstacle.}
\label{fig:indoor-terrain-setup}
\end{figure*}

\end{document}